\documentclass[11pt]{article}

\usepackage[preprint]{acl}
\usepackage{ragged2e}
\usepackage{times}
\usepackage{latexsym}

\newcommand{\sref}[1]{\S\ref{#1}}
\usepackage[T1]{fontenc}
\usepackage[utf8]{inputenc}
\usepackage{microtype}
\usepackage{inconsolata}
\usepackage{graphicx}
\usepackage{amsmath}
\usepackage{booktabs}
\usepackage{makecell}
\usepackage{multirow}
\usepackage{tabularx}
\usepackage[capitalise,nameinlink]{cleveref}

\title{More Computational Resources Do Not Ensure Higher Scholarly Impact: Evidence from Leading NLP Conference Papers}

\author{
  \textbf{Shuai Chen\textsuperscript},
  \textbf{Tong Bao\textsuperscript},
  \textbf{Jitong Peng\textsuperscript},
  \textbf{Chengzhi Zhang\textsuperscript{*}}
\\
\\
  \textsuperscript{}Department of Information Management, Nanjing University of Science and Technology, China\\
\\
  \small{
    \textbf{Correspondence:}
    \href{mailto:zhangcz@njust.edu.cn}{zhangcz@njust.edu.cn}
  }
}

\begin{document}
\maketitle
\begin{abstract}
Computational resources are increasingly central to NLP research, but how closely reported GPU capability aligns with scholarly impact remains unclear. We analyze 13,921 ACL, EMNLP, and NAACL main-conference papers published between 2020 and 2025, using GPU resources as our operational measure of computational resources. From full texts, we extract GPU models and counts, standardize each paper’s largest reported configuration into a comparable hardware-capability measure, and link these data to citation, award, topic, and institutional metadata. GPU reporting became more common but remained incomplete, while reported capability increased mainly through newer hardware generations and medium-scale multi-GPU configurations. Resource concentration substantially exceeded impact concentration: the annual top 20\% of GPU-quantifiable papers accounted for 83.9\%--89.9\% of reported GPU capability, but only 27\%--32\% of citations and 20\%--33\% of paper awards. In adjusted models, a tenfold increase in aggregate reported GPU capability was associated with a 3.52-percentage-point increase in within-NLP topic--year citation percentile, but increased model \(R^2\) by only 0.0042. GPU count showed more consistent positive associations with citation and award outcomes than newer hardware generation. Overall, reported GPU resources are associated with scholarly impact but provide little standalone explanation of research influence. \footnote{Code and data are available at \url{https://github.com/ChenShuai00/Computational-Resources}.}

\end{abstract}

\begin{figure}[t]
  \centering
  \includegraphics[width=\columnwidth]{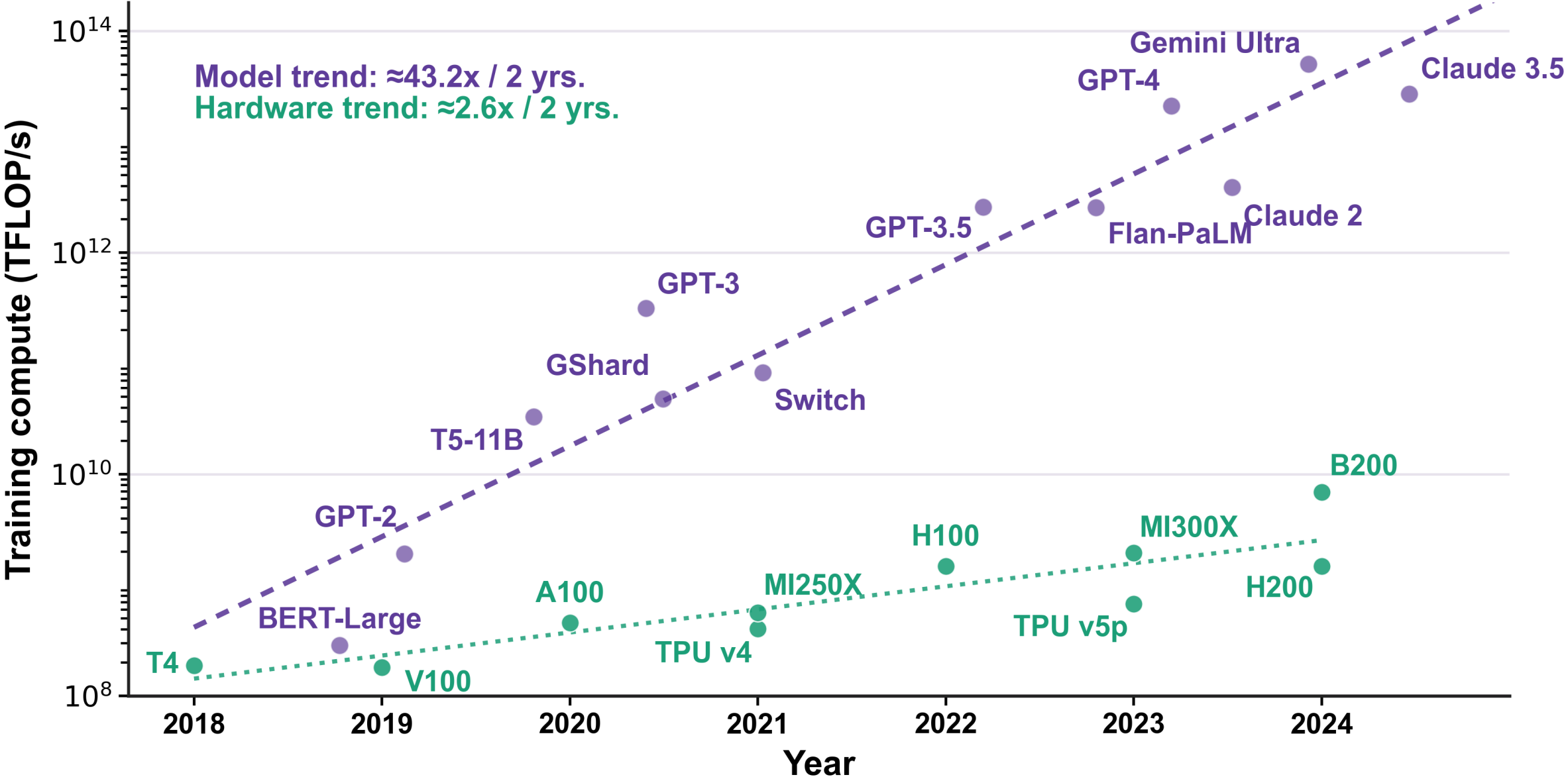}
  \caption{Estimated training compute for selected milestone language models has increased substantially over time, alongside advances in accelerator hardware.}
  \label{fig:training-compute}
\end{figure}

\section{Introduction}
The rapid scaling of language models has made computational resources, especially GPUs, increasingly central to NLP research \cite{schwartz2020green,sevilla2022compute,strubell2019energy}. Unequal access to compute may shape not only which research can be conducted, but also which work becomes visible, reusable, and influential \cite{storks2023nlp,besiroglu2024compute,hao2025role}. As shown in \cref{fig:training-compute}, the compute required to train milestone language models has grown far faster than the capability of individual accelerators, making access to larger and newer hardware configurations increasingly important. Yet it remains unclear whether differences in computational resources are systematically associated with scholarly impact across the broader NLP community. We therefore ask: \textbf{\textit{Are greater computational resources associated with higher scholarly impact, as reflected in citations and paper awards?}} 

Prior work documents unequal access to AI compute, including industry--academia asymmetries \cite{ahmed2020democratization}, advantages of elite institutions \cite{besiroglu2024compute}, and reporting and reproducibility burdens \cite{storks2023nlp}. Yet three issues remain unresolved: whether resource concentration is mirrored by impact concentration, whether associations persist after accounting for publication year, venue, and topic, and whether hardware deployment scale and hardware generation relate differently to citations and awards.

We analyze 13,921 ACL, EMNLP, and NAACL main-conference papers published between 2020 and 2025. In this study, we operationalize computational resources in terms of GPU resources. We extract and validate reported GPU models and counts, map them to standardized specifications, and define each paper's largest reported configuration as GPU count multiplied by theoretical peak Tensor FP16/BF16 throughput. This measure captures \emph{reported hardware capability}, not realized consumption: it excludes GPU hours, training FLOPs, cost, energy use, and utilization. Our analyses are therefore conditional on resources that are visible and standardizable from published papers.

We first describe GPU reporting and capability across time, topics, and institutional settings; then compare the concentration of capability, citations, and awards; and finally estimate covariate-adjusted associations. The primary citation outcome is the percentile within NLP topic--year cells, with citation models restricted to 2020--2023 to reduce window truncation. Complementary outcomes include the OpenAlex field-normalized percentile, raw citations, high-citation status, and awards. We further decompose aggregate capability into GPU count and hardware generation and test expanded author, team, and institutional controls.

The results show a \textbf{positive but limited alignment}. Reporting increased, and capability rose mainly through newer hardware generations and medium-scale multi-GPU configurations. During 2020--2023, the annual top 20\% of GPU-quantifiable papers accounted for 83.9\%--89.9\% of reported capability, but only 27\%--32\% of citations and 20\%--33\% of awards. High-capability papers were more likely to enter the citation top 10\%, yet most were not highly cited, and most highly cited papers lay outside the high-capability group.

Adjusted models reinforce this pattern. A tenfold increase in aggregate reported capability is associated with a 3.52-percentage-point increase in the primary citation percentile, but adds only 0.0042 to \(R^2\); the OpenAlex-normalized estimate is smaller and statistically imprecise. GPU count and newer hardware generation are both positively associated with the primary outcome, but GPU count is more consistent across complementary citation outcomes and is also positively associated with awards in linear-probability and Firth models. Aggregate capability and hardware generation show no comparably robust award evidence.

\begin{figure*}[t]
  \centering
  \includegraphics[width=1\textwidth]{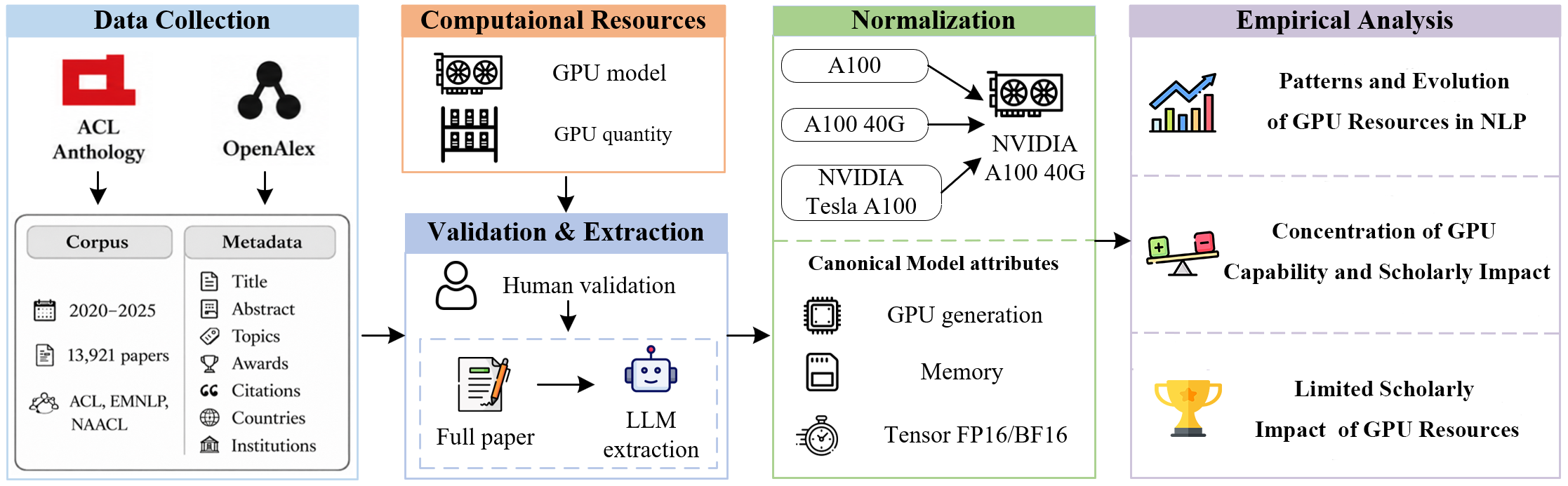}
  \caption{Overview of our computational resource measurement and analysis framework. We collect ACL, EMNLP, and NAACL papers from the ACL Anthology and enrich them with OpenAlex metadata, extract reported GPU models and quantities with human validation and LLM-based extraction, normalize raw hardware mentions to canonical GPU attributes, and use the resulting corpus to analyze reporting trends, reported compute scale, concentration across topics, organizations, and regions, and scholarly impact.}
\label{fig:framework}
\end{figure*}

Overall, reported GPU resources are positively associated with scholarly impact, but are neither necessary nor sufficient for high impact and explain little additional variation beyond observable research characteristics. \textbf{Our contributions are threefold:}  (1) We construct and validate a paper-level dataset of reported GPU configurations from 13,921 ACL, EMNLP, and NAACL main-conference papers; (2) We characterize temporal, topical, and institutional patterns in reported GPU capability and show that its concentration substantially exceeds that of citations and awards; (3) We provide field-normalized and covariate-adjusted estimates of compute--impact associations, showing limited incremental explanatory power and distinct patterns for GPU count and hardware generation.

\section{Related Work}
Computational resources have become an increasingly important component of AI research, prompting concerns about unequal access to compute and its implications for scientific participation. Prior studies have highlighted disparities between industry and academia, the concentration of computational resources within elite institutions, and the growing difficulty of reproducing compute-intensive experiments \cite{ahmed2020democratization,besiroglu2024compute,dodge2019show,storks2023nlp}. These works suggest that access to compute may influence who can conduct certain kinds of research and under what conditions. More recently, \citet{hao2025role} examined computing resources in foundation model research, finding associations between GPU resources and both publication outcomes and citation impact. However, how reported GPU resources are distributed and concentrated across NLP research, and whether differences in reported GPU capability are systematically associated with scholarly impact, remain underexplored.

\section{Methodology}
This section describes data collection (\sref{sec:data-collection}),
sample selection (\sref{sec:analysis-samples}), and GPU-capability
measurement (\sref{sec:measurement}). An overview of the framework is
shown in \cref{fig:framework}.

\subsection{Data Collection}
\label{sec:data-collection}

We collected papers published at three leading NLP conferences, ACL, EMNLP, and NAACL, from 2020 to 2025. After excluding papers without accessible PDFs, the final corpus comprised 13,921 papers. For each paper, we parsed the full text from PDFs using MinerU\footnote{\url{https://mineru.org.cn/}} \cite{wang2024mineru}, and collected the corresponding bibliographic metadata (authors, affiliations, citations, etc.) from OpenAlex\footnote{\url{https://openalex.org/}} \cite{priem2022openalex} via DOI, along with official conference records of paper awards and NLP topic labels classified by GPT-4o-mini \cite{openai2024gpt4omini}.  Details on data collection and preprocessing  are provided in \cref{app:data-collection}. 

\textbf{GPU Usage Extraction.} To determine GPU usage across the corpus, we first manually annotated 400 papers to create a human-validated evaluation set, including both GPU models and counts. To assess annotation reliability, two annotators independently annotated 120 overlapping papers after a pilot round and guideline refinement. Agreement was high: Cohen's $\kappa$ \cite{warrens2011chance} for identifying valid GPU-resource evidence was 0.94, with exact match rates of 90.83\% for GPU model and 87.50\% for GPU count. Disagreements were adjudicated by returning to the original evidence spans, and the remaining 280 papers were annotated by the primary annotator following the finalized guidelines. We then used LLMs to extract GPU information from the human-validated evaluation set and compared the results against the manual annotations. DeepSeek-v3.2 \cite{deepseekai2025deepseekv32} achieved a GPU-name F1 of 0.933 and an exact model-and-count F1 of 0.879, where the latter requires both the GPU model and the reported GPU count to match the human annotation. These results supported the reliability of the LLM-based extraction pipeline, which we subsequently applied to extract GPU usage records from the full corpus. Annotation guidelines and complete evaluation results are reported in \cref{app:annotation} to \cref{app:extraction}.

\textbf{GPU Normalization.} GPU mentions varied widely, including abbreviations, vendor names, memory variants, and generic descriptions. We normalized extracted hardware names using a conservative catalog-based procedure. Raw names were cleaned for case, vendor format, model string, memory notation, and uninformative suffixes, then mapped to a standard hardware catalog through exact matches, aliases, and rules for common GPU families and memory variants. Confirmed models missing from the catalog were added manually using vendor specifications and hardware documentation. Then, each standardized model was linked to memory capacity, GPU family, hardware generation, and theoretical peak Tensor FP16/BF16 throughput. Specifications were compiled from the Epoch AI Machine Learning Hardware dataset \cite{epochai2026hardware}, vendor sources, and manual verification. Full mapping rules and verification details are provided in \cref{app:Standardization}.

\textbf{Research Topic Controls.} Research topics may differ systematically in their computational requirements, making topic an important source of heterogeneity in reported GPU capacity. We therefore use topic only as a control and stratification variable. Because paper-level submission-area metadata are not consistently available across the 2020--2025 ACL, EMNLP, and NAACL proceedings, we assign each paper a single primary topic using a closed taxonomy of 29 categories adapted from the ACL Rolling Review (ARR) Area Keywords\footnote{\url{https://aclrollingreview.org/areas}}. The classification prompt is provided in \cref{app:prompts}.

\subsection{Analysis Sample Selection}
\label{sec:analysis-samples}
After extracting GPU resources information from 13,921 papers, we defined two analysis samples to address incomplete GPU reporting in NLP papers. The samples are based on whether a paper reports a standardizable GPU model and an explicit device count, as shown in \cref{tab:corpus-reporting-coverage}.

\textbf{Model-reported sample.} This group includes papers that report at least one standardized GPU model. It contains 6,900 papers, or 49.6\% of the full corpus. For papers that report a GPU model but not a GPU count, we set the count to one to construct a conservative lower bound on reported hardware capacity. This does not imply that the paper actually used only one GPU. Rather, it records the minimum capacity that can be confirmed from observable information. This sample is used mainly to measure GPU model disclosure and to support descriptive analyses.

\textbf{Strict sample.} This group includes papers that report both a standardized GPU model and an explicit GPU count. It contains 5,360 papers, or 38.5\% of the full corpus. Because both the GPU model and count are observable, reported GPU capacity can be computed directly at the paper-level. Regression analyses involving the intensity of computational capacity are therefore based primarily on the strict sample. 

\begin{table}[t]
\centering
\small
\setlength{\extrarowheight}{2pt}
\renewcommand{\arraystretch}{1.18}
\setlength{\tabcolsep}{4.5pt}

\begin{tabular}{
  >{\raggedright\arraybackslash}m{0.12\linewidth}
  >{\centering\arraybackslash}m{0.12\linewidth}
  >{\raggedright\arraybackslash}m{0.44\linewidth}
  >{\raggedright\arraybackslash}m{0.15\linewidth} 
}
\toprule
\textbf{Group} & \textbf{\#Papers} & \multicolumn{1}{c}{\textbf{Definition}} & \textbf{Example} \\
\midrule

Model-reported
& \shortstack{6,900\\(49.6\%)}
& At least one reported GPU model can be normalized to hardware specifications
& RTX 4090 \\

\addlinespace[4pt]

Strict
& \shortstack{5,360\\(38.5\%)}
& Both a standardizable GPU model and an explicit device quantity are observed
& 8×A100 GPU \\

\bottomrule
\end{tabular}
\caption{Dataset statistics used in this study.}
\vspace{-15pt}
\label{tab:corpus-reporting-coverage}
\end{table}

\subsection{GPU Capacity Estimation}
\label{sec:measurement}

We first map the GPU model to its theoretical peak Tensor FP16/BF16 throughput per card, measured in TFLOP/s. For paper \(i\), reported GPU capacity is defined as the largest observable GPU configuration reported in the paper:
\begin{equation}
\label{eq:gpu-capacity}
\mathit{GPU\_Capacity}_i
= \max_{g \in G_i} \left(n_{i,g} \times p_g\right)
\end{equation}
where \(G_i\) is the set of standardized GPU models reported in paper \(i\), \(n_{i,g}\) is the number of GPUs of model \(g\), and \(p_g\) is the theoretical peak Tensor FP16/BF16 throughput of one GPU of model \(g\). When a paper reports multiple GPU configurations, we use the largest configuration rather than summing across all configurations. This avoids double counting hardware used in different experiments, stages or runtime environments. 

\section{Results}

We organize the results as a sequential test of the paper's central
question. We first establish the empirical premise by showing how
reported GPU capability has changed and how unevenly it is distributed across NLP papers and research settings. We then test whether this concentration is mirrored in citations and awards. Finally, we estimate adjusted associations between reported GPU capability and scholarly
impact and examine whether the conclusions are robust across outcome
definitions, samples, and compute specifications.

\subsection{Patterns and Evolution of GPU Resources in NLP}
\label{sec:reporting}
Before examining whether reported GPU resources are associated with scholarly impact, we first establish how visible these resources are in the corpus and how their reported scale varies over time and across research contexts. 

\begin{figure}[t]
  \centering
  \includegraphics[width=\columnwidth]{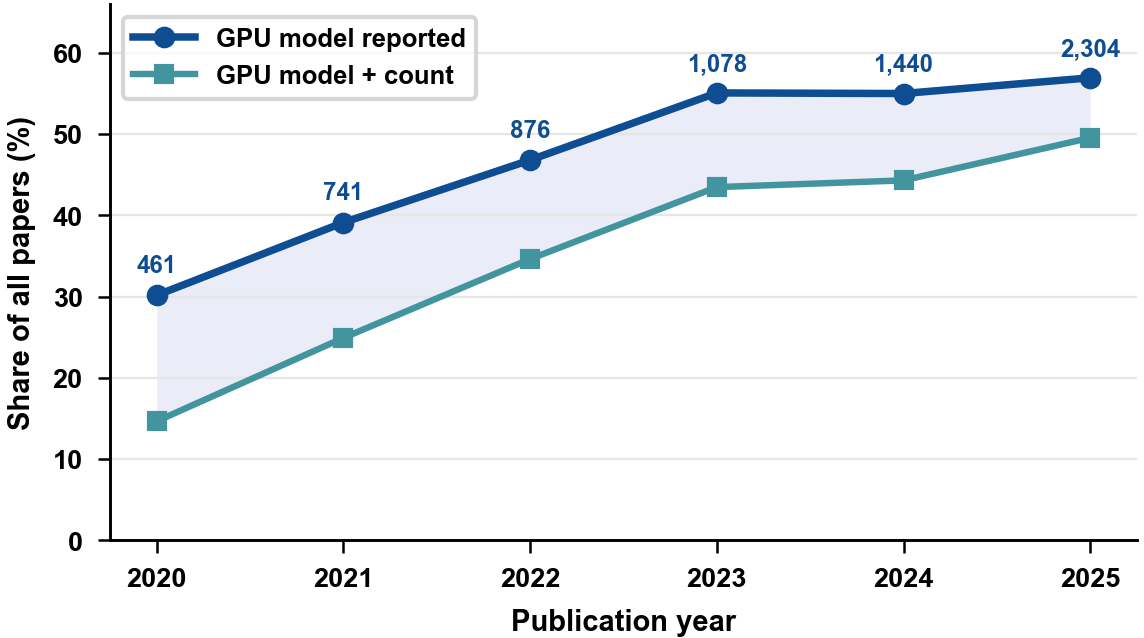}
  \caption{Reporting completeness of GPU resources in NLP papers.}
  \label{fig:reproting-completeness}
\end{figure}

\begin{figure*}[t]
  \centering
  \includegraphics[width=\textwidth]{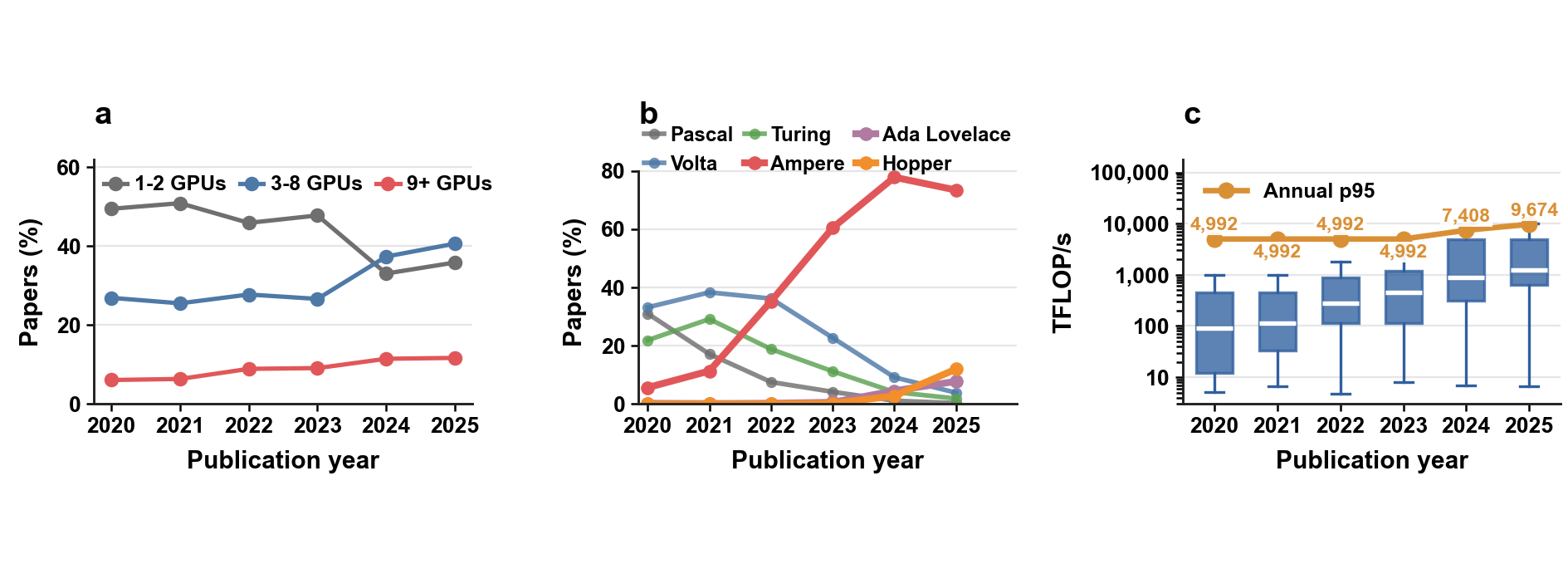}
  \vspace{-4em}
  \caption{Evolution of reported GPU scale, generation, and capacity in NLP research. The left panel shows GPU count over time; the middle panel shows the share of GPU generations; the right panel shows normalized reported GPU capacity distributions and annual 95th percentile (P95) values.}
  \label{fig:reported-GPU}
  \vspace{-0.4em}
\end{figure*}

\textbf{GPU Reporting Completeness and Trends.} GPU reporting became substantially more common between 2020 and 2025. As shown in \cref{fig:reproting-completeness}, the share of papers reporting at least one standardized GPU model increased from approximately 30\% in 2020 to 57\% in 2025. The share reporting both a GPU model and GPU count rose from approximately 15\% to 49\%. Thus, an increasing proportion of papers provides sufficient information to construct a paper-level measure of reported GPU capacity. Reporting nevertheless remains incomplete, and the following capacity analyses are therefore conditional on papers with positive, quantifiable GPU configurations. The missingness analyses in \cref{app:reporting-missingness} further show that, after controlling for year, venue, and research topic, organizational and collaboration characteristics add limited predictive information about whether a paper enters the reporting samples. This does not eliminate selection from incomplete reporting, but suggests that the observed organizational differences are unlikely to arise entirely from differences in reporting propensity.

\textbf{The Evolution of GPU Scale in NLP.} Among papers with quantifiable configurations, reported GPU capacity shifted substantially upward. As shown in \hyperref[fig:reported-GPU]{\cref*{fig:reported-GPU} (a)}, the share of papers reporting one or two GPUs declined from 49.5\% in 2020 to 35.9\% in 2025, while configurations using three to eight GPUs became more common. Configurations involving nine or more GPUs remained comparatively uncommon, accounting for 11.7\% of papers with reported GPU counts in 2025. At the same time, the reported hardware base moved toward newer GPU generations. V100-class hardware was most common in the earlier years, whereas A100-class GPUs became dominant from 2023 onward; H100-class GPUs began to appear in 2024 and 2025 but had not yet become dominant (\hyperref[fig:reported-GPU]{\cref*{fig:reported-GPU} (b)}). Correspondingly, median paper-level reported GPU capacity increased from approximately 91 TFLOP/s in 2020 to 1,248 TFLOP/s in 2025 (\hyperref[fig:reported-GPU]{\cref*{fig:reported-GPU} (c)}). Overall, the increase was driven mainly by the adoption of newer hardware and the expansion of medium-scale multi-GPU configurations, rather than by a field-wide transition to very large GPU clusters.

\begin{figure*}[t]
  \centering
  \includegraphics[width=0.8\textwidth]{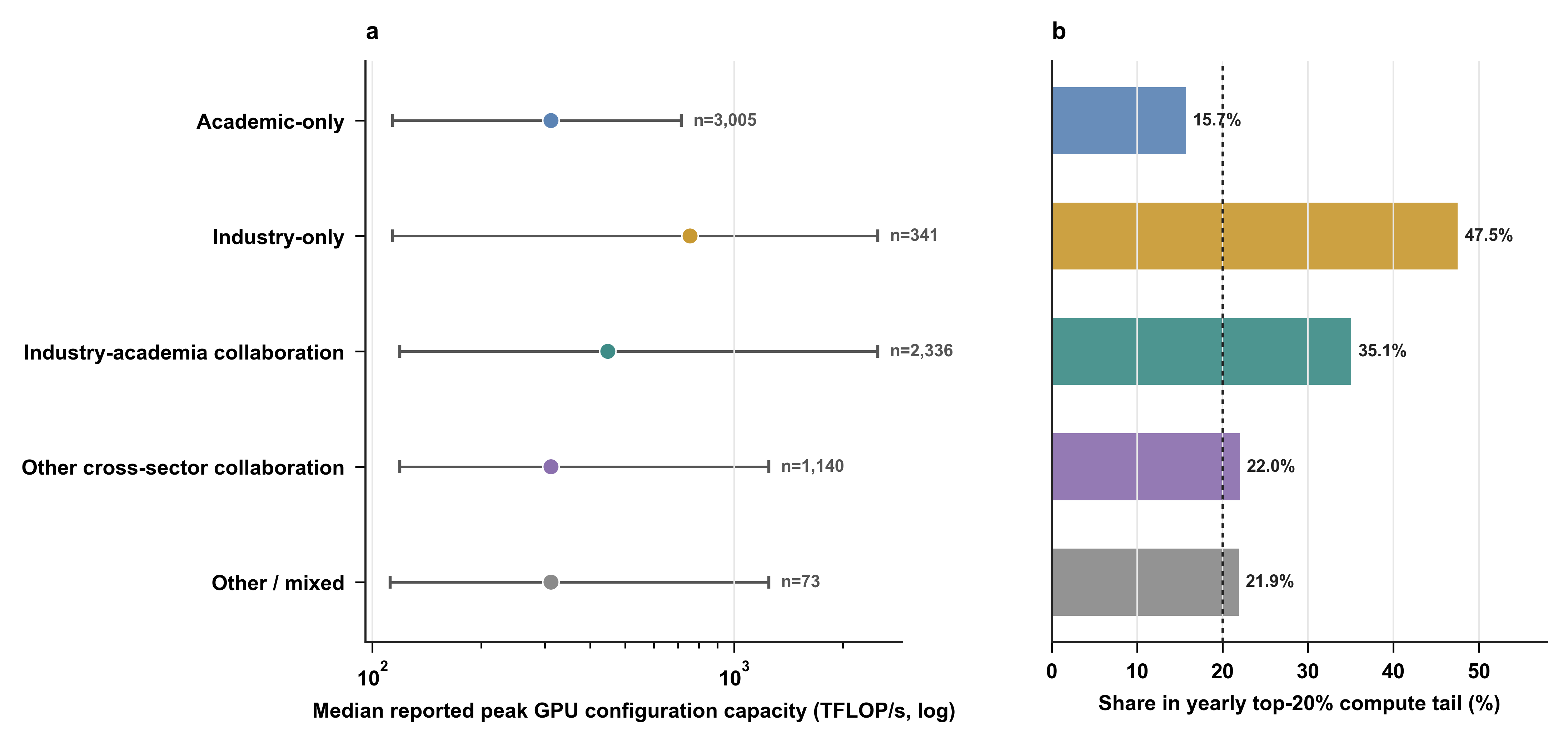}
  \caption{Reported GPU capacity by institutional access regime. 
Panel a shows median reported peak GPU configuration capacity across institutional access regimes. Points denote medians, horizontal intervals show the interquartile range, and sample sizes are reported on the right. 
Panel b shows share of papers in the yearly top-20\% reported GPU capacity tail. The dashed vertical line marks the 20\% reference level.}
\label{fig:reported-institution}
\end{figure*}

\textbf{Institutional Variation in GPU Capacity.}
The upward shift in reported capacity was accompanied by substantial heterogeneity across papers and institutional contexts. The paper-level distribution remained strongly right-skewed: most papers reported moderate configurations, while a relatively small group formed the high-capacity tail. Papers involving industry and collaborations between industry and academia reported higher median capacity and were more frequently represented in the annual high-capacity tail (\cref{fig:reported-institution}). 

Regression models controlling for year, venue, and research topic further support the concentration of reported capacity in industry-involved research (\cref{app:regression-institution}). The institutional indicators, however, overlap substantially: when they are entered jointly, industry participation remains strongly positive, whereas the collaboration coefficients are attenuated and should not be interpreted as independent institutional effects. Additional country- and region-level comparisons are reported in \cref{fig:gpu-country} of \cref{app:instifutional-correlate}.

\begin{figure}[t]
  \centering
  \includegraphics[width=\columnwidth]{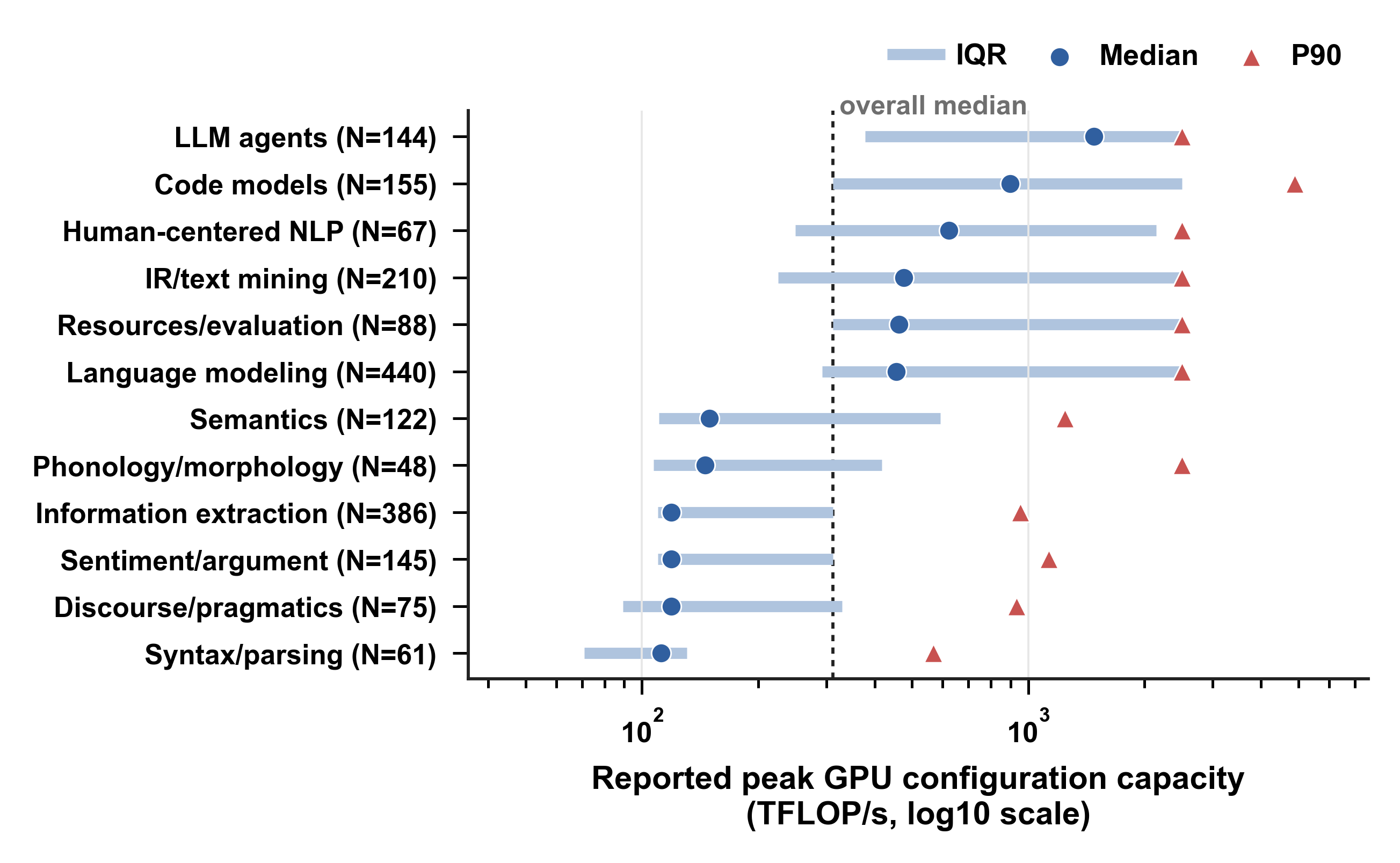}
  \caption{Reported GPU hardware capacity across major NLP topics: median, interquartile range (IQR), and the 90th percentile (P90). The full results for all 29 topic categories are reported in \cref{app:gpu-capacity-contexts} \cref{fig:topic}.}
  \vspace{-1em}
  \label{fig:topic-part}
\end{figure}

\textbf{GPU Capacity Across Research Topics.}
Heterogeneity in reported GPU capacity was also evident across research topics. As shown in \cref{fig:topic-part}, higher reported capacity was concentrated in topics closely connected to large-model development and deployment, including LLM agents, code models, and language modeling. More traditional areas, including syntax, sentiment analysis, and discourse and pragmatics, generally reported lower-capacity configurations. Because the figure presents only a subset of the 29 topic categories for readability, complete topic-level results are provided in \cref{fig:topic} of \cref{app:gpu-capacity-contexts}.

Together, these results establish the empirical setting for the impact analysis. Reported GPU resources became more visible and substantially more capable over time, but capacity remained concentrated in a relatively small upper tail and was systematically related to research topic and institutional context. Because these characteristics may also be associated with citations and awards, the following analyses first compare the concentration of reported GPU capacity with the concentration of scholarly outcomes and then estimate adjusted associations that account for these observed differences.

\subsection{Concentration of GPU Capability and Scholarly Impact} 
\label{sec:impact} 
The previous section documents a highly concentrated distribution of reported GPU capability across NLP papers. We now examine whether this concentration is mirrored in scholarly impact, including citations and paper awards. If reported GPU capability were closely aligned with impact, the most compute-intensive papers would also account for a comparably large share of citations and awards. 

\begin{figure*}[t] \centering \includegraphics[width=0.8\textwidth] {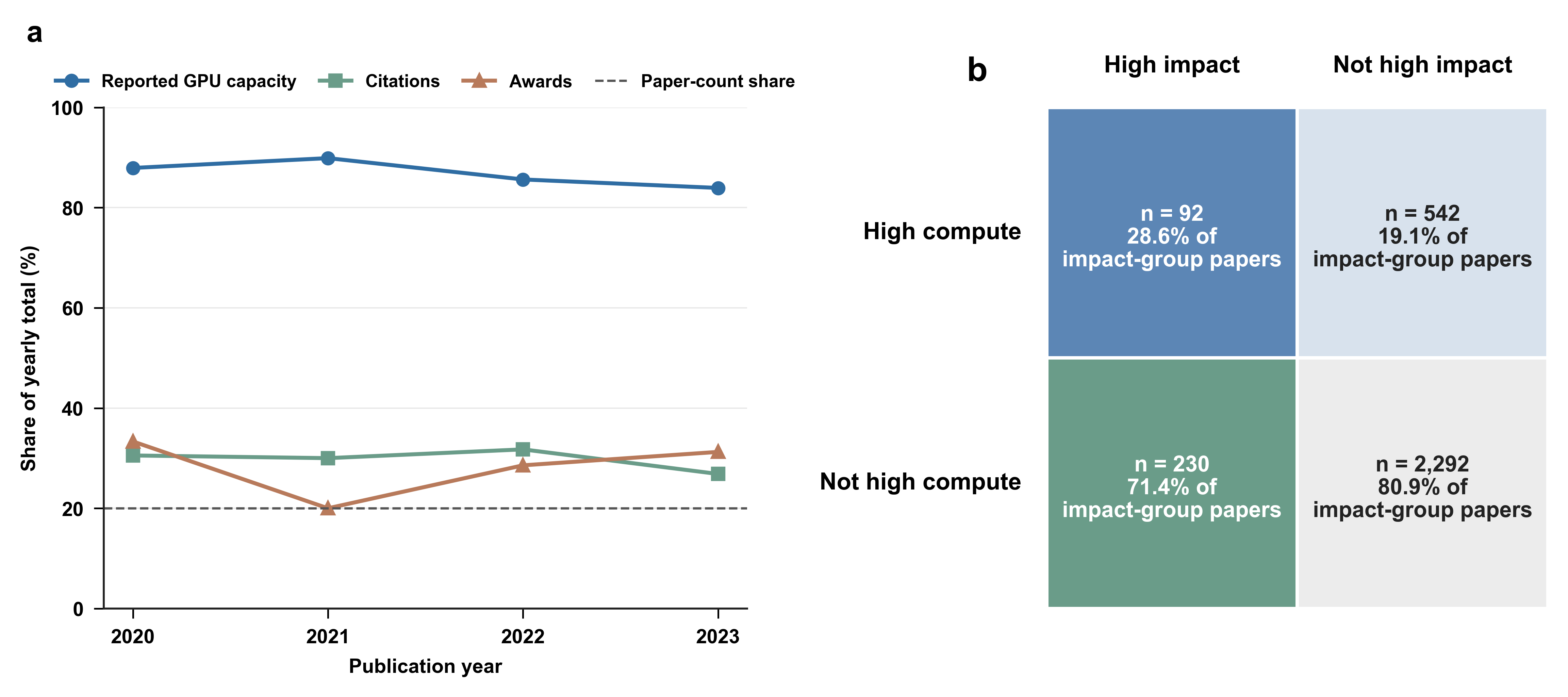} 
\caption{Concentration of reported GPU capability and scholarly impact. Panel (a) shows annual shares of total reported GPU capability, citations, and awards accounted for by papers in the top 20\% of reported GPU capability. Panel (b) shows overlap between high-capability and high-impact papers. High-capability papers are defined as the top 20\% by reported GPU capability, and high-impact papers as the top 10\% by citations within publication-year-by-venue groups.} 
\label{fig:impact-concentration} 
\end{figure*} 
For each year \(y\), let \(N_y\) denote the number of papers with positive reported GPU capability, and let
\begin{equation}
\label{eq:gpu-capacity-define}
C_{(1),y} \geq C_{(2),y} \geq \cdots \geq C_{(N_y),y}.
\end{equation}
denote capability in descending order. We define the high-capability group \(G_y\) as the top \(k_y=\lfloor 0.20N_y\rfloor\) papers. For any outcome \(Z\), the share attributed to this group is
\begin{equation}
\label{eq:gpu-capacity-share}
\mathit{S_y(Z)= \frac{\sum_{i\in G_y} Z_{iy}} {\sum_{i=1}^{N_y} Z_{iy}}}
\end{equation}
where \(Z\) is reported GPU capability, citations, or an award indicator. 

As shown in \hyperref[fig:impact-concentration]{\cref{fig:impact-concentration}(a)}, the top 20\% of papers account for 83.9\%--89.9\% of total reported GPU capability between 2020 and 2023. In contrast, the same set accounts for only 27\%--32\% of citations and 20\%--33\% of awards. Despite variation in award frequency, the gap between capability concentration and impact concentration remains substantial. 

We next examine overlap at the paper level. High-capability papers are defined as the top 20\% by reported GPU capability, and high-impact papers as the top 10\% by citations within publication-year-by-venue groups. As shown in \hyperref[fig:impact-concentration]{\cref{fig:impact-concentration}(b)}, 14.5\% of high-capability papers fall into the citation top 10\%, compared with 9.1\% among other papers. This corresponds to a 1.59× higher high-impact rate (a 59\% increase). The overlap is therefore positive but limited. Most high-capability papers (85.5\%) are not highly cited, and most highly cited papers fall outside the high-capability group. Reported GPU capability is thus neither necessary nor sufficient for high citation impact. 

We assess robustness by varying both thresholds: GPU-capability cutoffs at the top 10\%, 20\%, and 30\%, and citation cutoffs at the top 5\%, 10\%, and 20\%. \cref{tab:cutoff-sensitivity} reports the ratio of high-impact rates between papers above and below each GPU-capability cutoff. \begin{table}[t] \centering \small \begin{tabular}{lccc} \toprule \multirow{2}{*}{GPU-capability cutoff} & \multicolumn{3}{c}{Citation-impact cutoff} \\ \cmidrule(lr){2-4} & Top 5\% & Top 10\% & Top 20\% \\ \midrule Top 10\% & \(2.10\times\) & \(1.60\times\) & \(1.49\times\) \\ Top 20\% & \(2.14\times\) & \(\mathbf{1.59\times}\) & \(1.53\times\) \\ Top 30\% & \(1.93\times\) & \(1.55\times\) & \(1.49\times\) \\ \bottomrule \end{tabular} \caption{Robustness to alternative thresholds. Each entry reports the ratio of high-impact rates between papers above and below the GPU-capability cutoff. Values above one indicate higher citation impact among high-capability papers.} \label{tab:cutoff-sensitivity} \end{table} Across all specifications, the high-citation rate among high-capability papers is 1.49--2.14 times that among lower-capability papers. While the direction of association is stable, the magnitude of concentration in reported GPU capability is consistently stronger than that of citation impact. These results do not account for differences in topic, venue, team size, or organizational structure. The next section therefore estimates the relationship using continuous measures of reported GPU capability with covariate adjustment.

\begin{table*}[t] \centering \small \setlength{\tabcolsep}{3.5pt} \renewcommand{\arraystretch}{1.10} \begin{tabular*}{\textwidth} {@{\extracolsep{\fill}}lcccc@{}} \toprule Outcome & \shortstack{Aggregate GPU\\capability} & \shortstack{GPU\\count} & \shortstack{Ampere or\\newer} & \shortstack{\(\Delta R^2\)\\aggregate / joint} \\ \midrule NLP topic-year percentile (primary) & \(+3.52^{**}\) pp & \(+4.57^{***}\) pp & \(+4.16^{**}\) pp & \(0.0042 / 0.0093\) \\ OpenAlex percentile (secondary) & \(+1.26\) pp & \(+1.96\) pp & \(+2.02\) pp & \(0.0009 / 0.0033\) \\ \(\log(1+\mathrm{citations})\), OLS & \(+18.8^{***}\%\) & \(+24.7^{***}\%\) & \(+15.0^{*}\%\) & \(0.0052 / 0.0086\) \\ Citation count, PPML & \(+61.2^{***}\%\) & \(+69.3^{***}\%\) & \(+31.0^{*}\%\) & -- \\ Top-10\% cited, LPM & \(+3.84^{**}\) pp & \(+4.41^{**}\) pp & \(+2.17\) pp & \(0.0043 / 0.0051\) \\ Awarded, LPM & \(+0.86\) pp & \(+1.33^{*}\) pp & \(-0.25\) pp & \(0.0011 / 0.0018\) \\ \bottomrule \end{tabular*} \caption{Adjusted associations between reported GPU resources and scholarly impact.} \label{tab:adjusted-impact} \vspace{0.35em} \begin{minipage}{0.98\textwidth} \footnotesize \textit{Note.} Aggregate capability is measured as \(\log_{10}\) of the paper-level maximum reported GPU capability. GPU count and Ampere-or-newer are estimated jointly in a separate specification, without aggregate capability. Aggregate-capability and GPU-count effects are per tenfold increase; Ampere-or-newer effects are relative to earlier hardware. Effects are reported in percentage points (pp) for percentile and linear-probability outcomes and as percentage differences for log-citation OLS and PPML models. The final column reports incremental \(R^2\) for the aggregate and joint-component specifications. Citation models use \(N=2{,}194\); award models use \(N=5{,}357\). Full estimates and model statistics are reported in \cref{app:regression-Impact}. \(^{*}p<0.05\), \(^{**}p<0.01\), \(^{***}p<0.001\). \end{minipage} \end{table*}

\subsection{Limited Scholarly Impact of GPU Resources} 
\label{sec:adjusted-impact}

The distributional comparisons indicate that papers reporting greater GPU capability are more likely to attain high citation impact, but they do not account for differences in publication context, research topic,
or team structure. 

We next estimate associations between continuous measures of reported GPU resources and scholarly impact. Citation models use the strict 2020--2023 sample (\(N=2{,}194\)) to reduce citation-window truncation, whereas the award model uses the sample (\(N=5{,}357\)). All models include publication-year-by-venue fixed effects, primary-topic fixed effects, team-size controls, and organization-count controls. Our primary impact measure is the NLP topic-year citation percentile, which ranks each paper relative to papers published in the same year and assigned to the same primary NLP topic. The OpenAlex field-normalized citation percentile serves as a secondary normalized measure. We additionally examine \(\log(1+\mathrm{citations})\), raw citation counts, top-10\% citation status, and paper awards as complementary outcomes.

As shown in \cref{tab:adjusted-impact}, a tenfold increase in aggregate reported GPU capability is associated with a 3.52 percentage-point increase in the primary NLP topic-year citation percentile (\(95\%\ \mathrm{CI}=[1.27,5.77]\), \(p=0.002\)). However, adding reported GPU capability increases model \(R^2\) by only \(0.0042\). The association is smaller under the secondary OpenAlex field-normalized percentile: the estimated difference is 1.26 percentage points (\(95\%\ \mathrm{CI}=[-0.46,2.98]\), \(p=0.151\)), with an incremental \(R^2\) of \(0.0009\). Positive associations also appear in the count-based and high-citation
specifications. A tenfold increase in reported GPU capability is associated with 18.8\% higher \(1+\) citations, a 61.2\% increase in expected citation count in the PPML model, and a 3.84 percentage-point higher probability of belonging to the citation top 10\%. The corresponding aggregate association with paper awards is 0.86 percentage points and does not reach the conventional 0.05 threshold (\(p=0.056\)). To assess generalizability beyond main-conference papers, we replicate the citation analyses on Findings papers. Results are similar across tracks, with no significant slope differences and similarly modest incremental explanatory power (Appendix~\ref{app:findings-extension}).

We next distinguish the scale of GPU deployment from hardware
generation by entering reported GPU count and the Ampere-or-newer
indicator jointly. For the primary NLP topic--year percentile, a
tenfold increase in GPU count is associated with a 4.57
percentage-point difference
(\(95\%\ \mathrm{CI}=[1.96,7.18]\), \(p<0.001\)), while the use of
Ampere-or-newer hardware is associated with a 4.16 percentage-point
difference
(\(95\%\ \mathrm{CI}=[1.51,6.81]\), \(p=0.002\)), conditional on GPU
count. Under the OpenAlex percentile, the corresponding estimates are
approximately two percentage points but do not meet the 0.05 threshold
(\(p=0.052\) for GPU count and \(p=0.055\) for hardware generation).
Across the count-based and high-citation specifications, GPU count
shows the more consistent positive association, whereas newer hardware
generation is associated with citation intensity but not with top-10\%
citation status.

The award results are also dimension-specific. A tenfold increase in
GPU count is associated with a 1.33 percentage-point higher award
probability in the linear probability model, whereas the hardware-
generation coefficient is close to zero. This pattern is preserved in
a Firth rare-event model: a tenfold increase in GPU count is associated
with 1.71 times the odds of receiving an award
(\(95\%\ \mathrm{CI}=[1.19,2.44]\), Holm-adjusted \(p=0.0085\)),
whereas hardware generation remains statistically unsupported
(\cref{app:award-firth}). Because GPU count and hardware
generation are measured on different scales, their coefficient
magnitudes should not be interpreted as a direct ranking of their
importance. The component models instead show that the associations
vary across hardware dimensions and impact outcomes. Even under the
joint specification, incremental \(R^2\) remains below \(0.01\) for
every reported linear model.

Finally, we examine whether the primary result is explained by richer observable paper, author, and institutional characteristics. Using a common sample of 2,077 papers, the baseline association with the NLP topic-year percentile is 3.13 percentage points (\(95\%\ \mathrm{CI}=[0.81,5.45]\), \(p=0.008\)). Adding pre-publication measures of author citation history, team publication experience, institutional citation visibility, and collaboration structure reduces the estimate to 2.74 percentage points (\(95\%\ \mathrm{CI}=[0.37,5.12]\), \(p=0.024\)). A further specification including public-artifact availability yields a similar estimate of 2.65 percentage points (\(95\%\ \mathrm{CI}=[0.28,5.02]\), \(p=0.028\)). Across these specifications, the incremental \(R^2\) attributable to reported GPU capability declines from \(0.0032\) to \(0.0023\) and \(0.0022\), respectively. The OpenAlex percentile estimates remain small and statistically imprecise, while the log-citation coefficient is attenuated but remains positive. Full robustness results are reported in \cref{app:impact-expanded-controls}. 

Overall, reported GPU resources are positively associated with
citation impact under several specifications, including the primary
within-NLP percentile measure. The magnitude and precision of the
association nevertheless depend on the impact measure and
normalization reference set, and reported GPU resources add little
explanatory power beyond observable publication, topical, team, and
institutional characteristics. Award associations are dimension-
specific: GPU count is positively associated with award recognition
in both the linear probability and Firth rare-event models, whereas
aggregate GPU capability and hardware generation do not show
comparably robust evidence. These estimates should therefore be
interpreted as conditional associations rather than causal effects of
computational resources on scholarly impact.

\section{Discussion}

\paragraph{Positive but Limited Alignment with Impact.}
Greater reported capability is associated with several citation outcomes, and high-capability papers are more likely to be highly cited. However, the upper tail contains most reported GPU capability but a much smaller share of citations and awards; most high-capability papers are not highly cited, and most highly cited papers lie outside that tail. Aggregate capability also adds little incremental explanatory power, and its OpenAlex field-normalized estimate is small and statistically imprecise. Reported GPU resources may expand experimental possibilities, but they neither ensure nor are necessary for scholarly impact.

\paragraph{Resource and Impact Concentration Are Not Equivalent.}
Reported capability is concentrated in large-model topics and industry-involved research. This uneven distribution matters because resource availability may shape which experiments are feasible, even though it does not determine which work becomes influential. Broader access to infrastructure and evaluation based on scientific contribution are therefore complementary goals: reducing resource barriers supports participation, while scholarly contribution should not be inferred from hardware scale. Better reporting of GPU models, counts, runtime, utilization, and externally provided compute would help future studies distinguish available capability from actual consumption.

\paragraph{API-Mediated Compute.} The growing use of LLM APIs shifts the relevant constraint from researcher-owned GPUs to platform-mediated model access. API-only studies may rely on substantial upstream compute while reporting no local hardware, making this dependence largely invisible to our GPU-configuration measure; they should therefore be understood as relying on externally mediated compute rather than as compute-free research. Because we do not treat non-reporting as zero capability, this issue primarily limits the coverage and interpretation of our measure, although it may understate compute dependence in application-oriented research and reshape the observed association with scholarly impact. This concern is more consequential for recent descriptive trends than for our main citation models, which use the 2020--2023 sample. More broadly, the emerging resource divide increasingly concerns access to models, interfaces, budgets, and platform transparency, in addition to GPU ownership.

\section{Conclusion}

We analyze reported GPU resources in 13,921 ACL, EMNLP, and NAACL main-conference papers published between 2020 and 2025. GPU reporting became more common but remained incomplete, while reported capability increased mainly through newer hardware generations and medium-scale multi-GPU configurations. Reported GPU capability was concentrated in a small upper tail and varied systematically across topics and institutional settings, yet its concentration far exceeded that of citations and awards. Adjusted models showed positive but outcome-dependent associations: aggregate capability was associated with several citation measures but added little explanatory power, whereas GPU count showed the most consistent pattern across citation outcomes and the only award association supported by both linear-probability and rare-event models. Overall, reported GPU resources are important infrastructure for contemporary NLP research, but they do not ensure scholarly impact and provide only a limited standalone explanation of research influence.

\section*{Limitations}

We acknowledge several limitations in our work.

\textbf{Reported Resources and Sample Selection.}
Our data record only standardizable GPU models and counts explicitly reported in papers. Nonreporting papers may still use local hardware, while API-based or managed services may involve substantial remote computation without revealing the underlying GPUs. In a manual audit, only 92 of 240 GPU-reporting papers (38.3\%) contained a consumption-related signal, and those signals were too heterogeneous for a comparable measure (\cref{app:compute_audit}). The analyses are therefore conditional on visible, extractable, and standardizable configurations; incomplete reporting may introduce selection that the missingness checks cannot fully remove.

\textbf{Hardware-Capability Measurement.}
Aggregate reported GPU capability is based on the largest observed configuration and theoretical peak Tensor FP16/BF16 throughput. It does not measure runtime, utilization, memory bandwidth, interconnect performance, software efficiency, hyperparameter search, or cumulative use across experiments. Identical configurations can represent anything from short inference runs to prolonged training. Although the extraction and normalization pipeline achieved high validation performance, residual errors in model identification, counts, and hardware mapping may remain.

\textbf{Impact Measures and Observational Design.}
Citations, high-citation status, and paper awards are incomplete proxies for scholarly value. Restricting citation models to 2020--2023 and using within-topic--year percentiles reduces citation-window and field differences but does not eliminate them; the smaller OpenAlex field-normalized estimates also show sensitivity to the reference set. Topic normalization relies on a single assigned primary topic, and unobserved author, institutional, and project characteristics may remain correlated with both reported resources and impact. The estimates are therefore conditional associations and should not be interpreted causally.

\textbf{Corpus and Metadata Scope.}
The corpus covers only ACL, EMNLP, and NAACL main-conference papers from 2020 to 2025. The findings may not generalize to workshops, journals, arXiv preprints, industrial technical reports, or other NLP and machine-learning venues. Organizational and geographic analyses rely on affiliation metadata and full-counting rules, which cannot identify resource ownership, researcher mobility, or access to shared and cross-national infrastructure; these comparisons should be interpreted descriptively.

\section*{Ethics Statement}

\textbf{Data and Privacy.}
This study uses publicly available scholarly articles and metadata from
the ACL Anthology \cite{bollmann-etal-2023-two} and OpenAlex
\cite{priem2022openalex}, supplemented by official award records and
public hardware specifications. Author names and affiliations were used
only for bibliographic linkage and aggregate analysis. We did not
collect private communications, reviewer information, non-public
contact information, or protected demographic attributes. Geographic
variables refer to institutional locations rather than authors'
nationality or ethnicity.

\textbf{Automated Processing and Validation.}
Only publicly released paper text was processed using MinerU and
LLM-based systems for GPU-resource extraction and topic classification;
no confidential submissions or peer-review materials were provided to
these systems. The GPU extraction pipeline was evaluated against human
annotations to assess automated-processing errors.

\textbf{Use and Release.}
Our measures represent reported GPU configurations rather than actual
compute consumption, resource ownership, researcher ability, or
scientific quality. Institutional and geographic comparisons are
descriptive, and the observed associations should not be interpreted
causally or used for individual or institutional evaluation. We release
code and derived data subject to the licenses and usage requirements of
the original sources, without redistributing source full texts or
unnecessary direct identifiers.

\section{Acknowledgements}

This paper was supported by the National Natural Science Foundation of China (Grant No.72074113) and 2026 Special Project of the Innovation Intelligence Professional Committee of the Chinese Society for Scientific and Technical Information (CSSTI).

\bibliography{custom}

\appendix
\section{Additional Related Work}
\label{app:related-work}
\textbf{Compute infrastructure in scaling NLP models.} As Transformers, pretrained language models, large language models, code models, multimodal systems and agents have developed, computational resources have become core infrastructure for NLP research. Prior work shows that increases in model size, data scale and training compute can drive sustained improvements in model performance. The Transformer architecture provided a foundation for large-scale sequence modeling, while pretrained models such as BERT, T5 and the GPT series shifted NLP from models for specific tasks toward large-scale pretraining \cite{vaswani2017attention,devlin2019bert,brown2020language,raffel2020exploring}. Research on scaling laws further shows that performance can improve in relatively regular ways as parameter counts, data and training compute increase \cite{kaplan2020scaling,hoffmann2022empirical}. More recent dense models and sparsely activated models have reinforced the connection between model capability and computational infrastructure  \cite{lepikhin2020gshard,fedus2022switch,smith2022using,thoppilan2022lamda,chowdhery2023palm}.

\textbf{Transparency in reporting compute for paper experiments.} As NLP models have become more dependent on hardware resources, researchers have paid growing attention to computational cost, energy use, carbon emissions and reproducibility. Work on Green AI and carbon accounting argues that model performance should not be evaluated apart from training cost, energy use and environmental impact \cite{strubell2019energy,schwartz2020green,lacoste2019quantifying,henderson2020towards,patterson2021carbon,luccioni2023estimating,luccioni2024power}. Research on reproducibility and documentation similarly emphasizes the need to report development budgets, hyperparameter search, random seeds, computational infrastructure and data documentation, so that results can be compared, verified and reproduced \cite{dodge2019show,dodge2020fine,dror-etal-2018-hitchhikers,sevilla2022compute}. Work on efficient NLP further shows that lowering computational barriers matters for broader research participation and deployment \cite{treviso2023efficient,storks2023nlp}. 

\textbf{Stratification in access to compute.} Beyond technical cost and reporting transparency, access to computational resources may shape the distribution of opportunities in AI research. When frontier model training, large-scale fine tuning and complex system evaluation require expensive hardware, researchers without sufficient infrastructure may face structural disadvantages. Prior work suggests that AI research opportunities, institutional visibility and scholarly influence may increasingly concentrate among organizations with proprietary or large-scale compute resources \cite{ahmed2020democratization,besiroglu2024compute}. This concern is also connected to cumulative advantage and the Matthew effect, in which actors who already hold resource advantages are more likely to gain attention, reputation and further resources \cite{merton1968matthew,merton1988matthew,bol2018matthew}. 

\begin{table*}[t]
\centering
\small
\begin{tabular}{p{0.25\textwidth} p{0.68\textwidth}}
\toprule
\textbf{Type of dataset} & \textbf{Role in analysis} \\
\midrule
Paper corpus & Defines the 13,921 ACL/EMNLP/NAACL main conference papers. \\
Parsed paper text & Source for GPU hardware mentions and evidence spans. \\
OpenAlex metadata & Provides citation counts, author/team size, and citation percentile measures. \\
Organization variables & Capture industry participation, industry-academia collaboration, cross-sector collaboration, international collaboration, and organization counts. \\
Topic labels & Provide a 29-topic taxonomy based on title and abstract classification. \\
Award labels & Identify best paper and related award indicators for 2020--2025 main conference papers. \\
\bottomrule
\end{tabular}
\caption{Dataset components.}
\label{tab:dataset_components}
\end{table*}

\section{Data and Preprocessing}
\label{app:data-preprocessing}

\subsection{Data Collection}
\label{app:data-collection}
\textbf{Raw Data.} The corpus consists of ACL, EMNLP, and NAACL leading conference papers from 2020 to 2025. We use 2020 as the starting point because it captures the post-BERT era of large-scale pretrained language models, when GPU-based computation became increasingly central to NLP and reproducibility checklists made reporting practices more comparable across these conferences \cite{raffel2020exploring,brown2020language,Magnusson2023ReproducibilityIN}. After excluding papers without usable PDFs or core metadata, the final corpus contains 13,921 papers. For each paper, we recorded year, venue, title, abstract, authors, affiliations, countries and regions, citation measures, award labels, and topic labels. PDFs were parsed with the MinerU API \cite{wang2024mineru} to obtain full text for GPU evidence extraction. 

OpenAlex \cite{priem2022openalex} was used to supplement citation counts, team size, normalized citation percentiles, and selected author and institution metadata. Although OpenAlex provides broad, open, and reproducible bibliographic coverage, its citation and affiliation metadata are database-dependent and may contain omissions or disambiguation errors. We therefore use it as a supplementary metadata source, combine it with official conference records where available, and report the retrieval date for reproducibility.

\textbf{Affiliation and Organization Metadata.} Affiliation data describe the organizations represented by paper authors and the types of those organizations. We first parsed author institutions from the affiliation information reported in the papers, and supplemented or disambiguated these records using OpenAlex institution metadata when needed. We then mapped raw institution names to normalized organization records. Each organization was assigned an organization type and a country code. Organization types include higher education or research institutions, industries, government agencies, nonprofit organizations, medical institutions, research facilities, archives, and other organizations. Using these normalized records, we constructed variables at the paper-level. These variables measure the number of participating organizations, the number of participating countries, industry participation, academic participation, collaboration between industry and academia, collaboration across sectors and international collaboration. Country and region variables were derived from the country codes of participating institutions. In geographic analyses, a country receives one paper count if at least one institution from that country participated in the paper.

\textbf{Award Metadata.} Award labels identify whether a paper received a main conference paper award from ACL, EMNLP or NAACL. We collected official award records for the target years and venues. Awarded papers were matched to the main corpus using ACL Anthology \cite{bollmann-etal-2023-two} paper IDs. A paper was labelled as awarded if it appeared in the relevant best paper, outstanding paper, honorable mention or other main conference award list. Otherwise, it was labelled as not awarded. This variable is used to examine whether reported GPU hardware capacity is associated with community-level paper recognition. Because awards are rare, award-based analyses are treated as a supplementary measure of scholarly impact. They are not interpreted as a complete measure of paper quality.

\textbf{Research Topic Metadata.}
Research topic labels describe the primary research area of each paper. Using each paper's title and abstract, we assigned one primary NLP topic from a taxonomy of 29 topics derived from the ACL Rolling Review (ARR) area keywords\footnote{https://aclrollingreview.org/areas}. The classification focuses on the paper's core research question and main contribution, rather than incidental methods, datasets, models, or tools. For example, a paper that uses a large language model for sentiment analysis is classified by its main task, such as sentiment analysis, rather than as a large language model topic solely because it uses such a model.

We used this ARR taxonomy rather than unsupervised topic models such as BERTopic \cite{grootendorst2022bertopic} because our goal was not to discover latent themes, but to assign papers to stable, interpretable categories consistent with field practice. This design supports comparisons across years and venues, as well as regression analyses. Unsupervised topic models are useful for exploratory analysis, but their clusters can vary with embedding models, hyperparameters, corpus composition, and subsequent topic naming, which may reduce comparability across analyses. Topic labels are used to compare reported GPU capacity across research areas and as topic controls in the regression models. \cref{app:prompts} provides the GPT-4o-mini prompt used for NLP topic classification.

All computational resource measures are constructed at the paper-level. Organization and geographic analyses expand paper-level observations to the organization or country level using affiliation information. Because hardware information is observed only through explicit reporting in paper text, all downstream analyses are conditional on whether a paper reports identifiable and standardizable GPU hardware. \cref{tab:dataset_components} summarizes the dataset components used in the analysis.

\begin{table*}[t]
\centering
\small
\begin{tabular}{
  >{\raggedright\arraybackslash}p{0.20\textwidth}
  >{\raggedright\arraybackslash}p{0.30\textwidth}
  >{\raggedright\arraybackslash}p{0.43\textwidth}
}
\toprule
\textbf{Case type} & \textbf{Example pattern} & \textbf{Annotation decision} \\
\midrule
Complete GPU reporting
& \emph{trained on 8 NVIDIA A100 80~GB GPUs}
& Annotate GPU model, memory, and quantity; eligible for paper-level capacity estimation. \\

Model-only reporting
& \emph{experiments ran on A100 GPUs}
& Annotate GPU model; quantity is \texttt{null}, and lower-bound analyses assign one GPU. \\

Ambiguous hardware reporting
& \emph{experiments were run on GPU servers}
& Not standardizable to a specific GPU model; excluded from capacity estimation. \\

Citation or prior-work hardware
& \emph{Brown et al. used 1024 TPU v3 chips}
& Prior-work hardware; not annotated as this paper's hardware. \\

Model/API/software names
& \emph{we use GPT-4o, Claude, Gemini, and vLLM}
& Model/API/software names are not hardware; not annotated. \\
\bottomrule
\end{tabular}
\caption{Boundary cases in computational resource annotation.}
\label{tab:boundary_cases}
\end{table*}

\subsection{Annotation Guidelines and Boundary Cases}
\label{app:annotation}
To evaluate the automatic extraction pipeline, we constructed a human-validated annotation set. The validation set was primarily sampled from papers whose EMNLP 2025 Responsible NLP Checklist responses\footnote{https://aclanthology.org/attachments/2025.emnlp-main.1.checklist.pdf} to C1 indicated that model parameters, computational budget, or computing infrastructure were reported. Because C1 responses often point to the relevant paper sections, we used these locations as high-recall candidate cues rather than treating the checklist answers themselves as GPU-resource labels.

\textbf{Annotation Details.} The annotation set contains 400 papers. One primary annotator, an author of this paper with two years of NLP research experience, annotated all 400 papers. To assess inter-annotator agreement, a second doctoral student with two years of related research experience independently annotated a random subset of 120 papers. Annotators judged whether each candidate passage contained evidence of GPU resources actually used in the current paper's experiments. When such evidence was present, they recorded the GPU model, GPU count, memory configuration, and the corresponding evidence span. All annotations were conducted in Label Studio \cite{labelstudio2020}.

\textbf{Annotation Results.} Before independent annotation, the two annotators conducted a pilot round, discussed ambiguous cases, and revised the annotation guidelines. After the guidelines were fixed, they independently annotated the 120 overlapping papers. Agreement was high: Cohen's $\kappa$ \cite{warrens2011chance} for the binary label of whether valid GPU-resource evidence was present was 0.9409, the exact match rate for GPU model was 90.83\%, and the exact match rate for GPU count was 87.50\%. Disagreements in the overlapping subset were adjudicated by returning to the original evidence spans and resolving them through discussion. The remaining 280 papers were annotated by the primary annotator following the finalized guidelines.

\begin{table*}[t]
\centering
\small
\setlength{\tabcolsep}{8pt}
\renewcommand{\arraystretch}{1.12}
\begin{tabular}{llccc}
\toprule
\textbf{Model} & \textbf{Matching criterion} & \textbf{Precision} & \textbf{Recall} & \textbf{F1} \\
\midrule
\multirow{2}{*}{\texttt{deepseek-v3.2}}
& Name only & \(0.918 \pm 0.0028\) & \(\mathbf{0.948 \pm 0.0013}\) & \(\mathbf{0.933 \pm 0.0019}\) \\
& Exact match & \(0.865 \pm 0.0021\) & \(\mathbf{0.894 \pm 0.0024}\) & \(0.879 \pm 0.0020\) \\
\midrule
\multirow{2}{*}{\texttt{gpt-4o-mini}}
& Name only & \(0.907 \pm 0.0099\) & \(0.842 \pm 0.0066\) & \(0.874 \pm 0.0069\) \\
& Exact match & \(0.864 \pm 0.0130\) & \(0.802 \pm 0.0115\) & \(0.832 \pm 0.0115\) \\
\midrule
\multirow{2}{*}{\texttt{gpt-5.4-mini}}
& Name only & \(\mathbf{0.927 \pm 0.0092}\) & \(0.933 \pm 0.0096\) & \(0.930 \pm 0.0092\) \\
& Exact match & \(0.842 \pm 0.0126\) & \(0.848 \pm 0.0133\) & \(0.845 \pm 0.0128\) \\
\midrule
\multirow{2}{*}{\texttt{gemini-3-flash-preview}}
& Name only & \(0.924 \pm 0.0031\) & \(0.940 \pm 0.0045\) & \(0.932 \pm 0.0037\) \\
& Exact match & \(\mathbf{0.876 \pm 0.0046}\) & \(0.890 \pm 0.0043\) & \(\mathbf{0.882 \pm 0.0043}\) \\
\bottomrule
\end{tabular}
\caption{LLM extraction performance on the human-validated set. Scores are reported as mean \(\pm\) standard deviation. Bold values indicate the best score within each matching criterion.}
\label{tab:llm_extraction_performance}
\end{table*}

\textbf{Annotation Guidelines}
\label{app:annotation-guidelines}
The annotation task was to identify accelerated computing hardware that authors explicitly reported as being used for the focal paper’s experiments. Annotators first judged whether a candidate text passage contained valid evidence of computational resources. If valid evidence was present, they recorded the hardware name, hardware count, memory information when explicitly stated, and the corresponding evidence span. Annotation was limited to hardware information that appeared directly in the text. Annotators were not allowed to infer hardware counts or memory capacity from model size, common cloud configurations or default hardware specifications.

Included hardware types were GPUs, TPUs, NPUs, IPUs and similar AI accelerators. Excluded items were CPUs, system memory, storage, runtime, GPU hours, cloud service costs, carbon emissions, model names, LLM API names, software frameworks, algorithms and optimizers. Hardware mentions were also excluded if they appeared only in cited papers, prior work, external baselines or hypothetical settings. When a hardware name was explicitly reported, annotators preserved the manufacturer, model and memory configuration. For example, “4 NVIDIA A100 GPUs, each with 80 GB” was recorded as hardware name “NVIDIA A100 80GB” and count 4. Minor format normalization was allowed, such as rewriting “V100 32G” as “V100 32G” in a consistent format. However, annotators were not allowed to fill missing information using external knowledge. If the text explicitly stated that a single device was used, the count was recorded as 1. If the count was not stated, the count was recorded as null. \cref{tab:boundary_cases} shows Boundary-cases in computational resource annotation.

For baseline models, annotators distinguished hardware used in the original training of an external baseline from hardware used by the authors of the focal paper. If a paper only described the hardware used to train a baseline model in prior work, that hardware was not annotated. If the paper explicitly stated that the authors reproduced, trained, fine-tuned or evaluated a baseline model on a specific GPU configuration, that hardware was annotated. When a sentence mentioned both prior work hardware and hardware used in the focal paper, annotators recorded only the hardware used by the authors of the focal paper.

\subsection{Extraction and Topic Classification Prompts}
\label{app:prompts}
\begin{figure*}[t]
  \centering
  \includegraphics[width=0.95\textwidth]{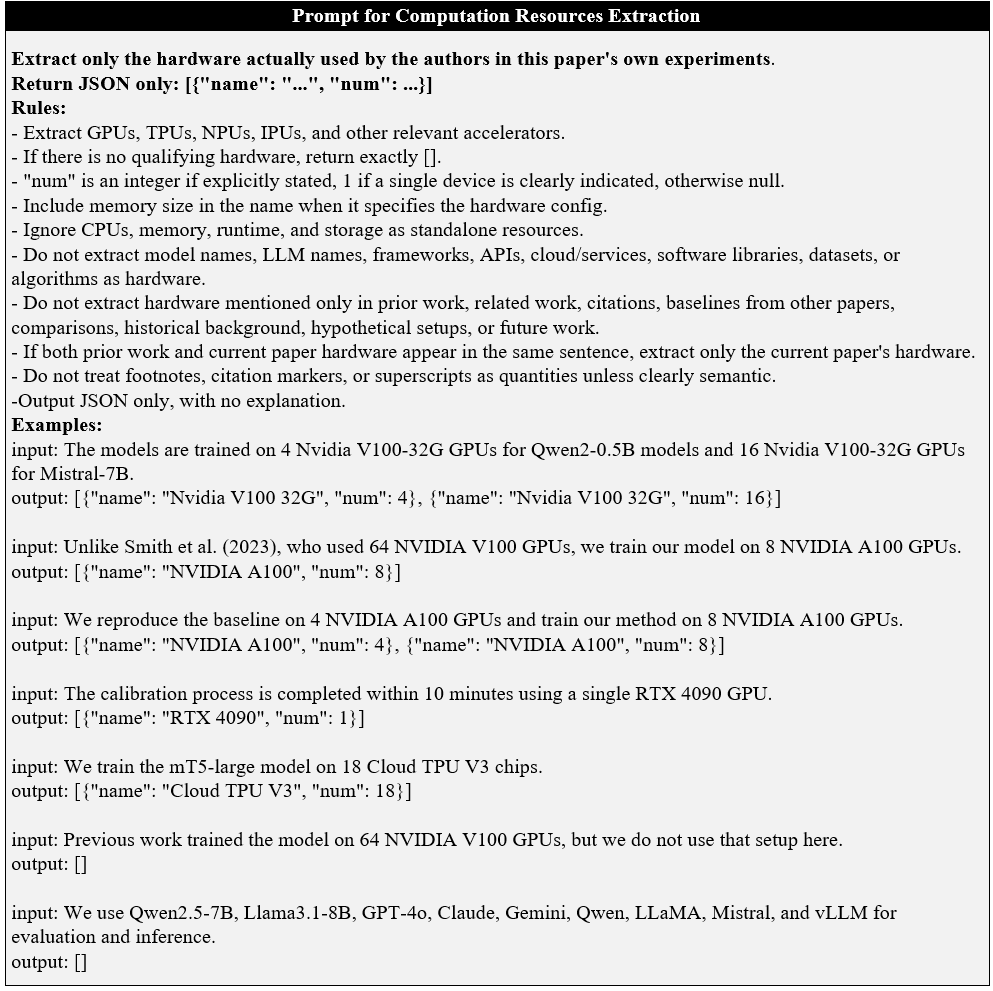}
  \caption{Prompt used to extract computational resources.}
  \label{fig:prompt-extract}
\end{figure*}

\begin{figure*}[t]
  \centering
  \includegraphics[width=0.95\textwidth]{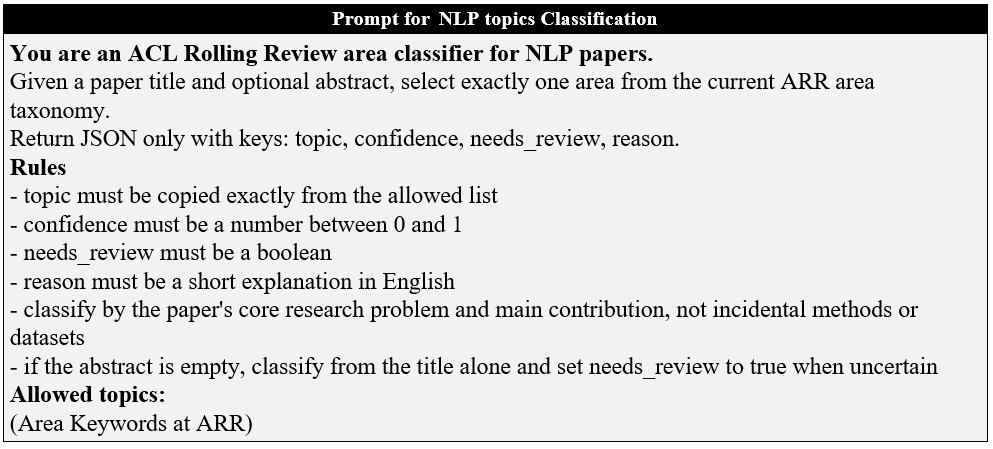}
  \caption{Prompt used to classify NLP topics.}
  \label{fig:prompt-classify}
\end{figure*}

The GPU extraction prompt instructed the model to identify only hardware reported as being used in the focal paper. The model returned GPU model names and counts in structured fields. The prompt explicitly prohibited inference of missing information. It also instructed the model not to treat model names, API names or software names as hardware. Hardware mentioned in related work, cited papers or external baselines was not to be treated as a resource used by the focal paper. \cref{fig:prompt-extract} shows the prompt to extract computational resources. \cref{fig:prompt-classify} shows the prompt to classify NLP research topics.

\subsection{Extraction Evaluation and Large-Scale Extraction}
\label{app:extraction}

\textbf{Extraction Evaluation.} We evaluated four models on the computational resource extraction task using the human-validated evaluation set. The four models are DeepSeek-V3.2 \cite{deepseekai2025deepseekv32}, GPT-4o-mini \cite{openai2024gpt4omini}, GPT-5.4-mini \cite{openai2026gpt54mini}, and Gemini-3-Flash-Preview \cite{google2026gemini3flashpreview}. The evaluation considered two capabilities: identifying GPU model names, and jointly extracting GPU model names and GPU counts. All models used the same candidate text passages, field definitions, output format and default parameter settings. Each model was run independently five times, and we report the mean and standard deviation. \cref{tab:llm_extraction_performance} reports model performance under these two settings. The name only setting evaluates extraction of GPU model names alone, whereas the exact match setting requires both the GPU model and the GPU count to be correct.

Overall, Gemini-3-Flash-Preview achieved the best performance under the exact-match criterion, with an F1 score of 0.882, while DeepSeek-V3.2 performed only slightly lower, with an F1 score of 0.879. After balancing model performance against the cost of processing approximately 274 million input tokens in the full sample, we used DeepSeek-V3.2 to extract GPU models and quantities from full-text candidate passages.

\textbf{Large-Scale Extraction.} For the large-scale extraction stage, we used DeepSeek V3.2 to extract reported GPU resource information from 13,921 ACL, EMNLP and NAACL main conference papers published between 2020 and 2025. Extraction was based on full paper text, but we first removed the introduction and related work sections. This reduced the risk that the model would mistake hardware mentioned in background discussion, prior work or cited papers for resources used in the focal paper. We then split each paper into text chunks of 3,000 characters with a 300-character overlap.

This setting preserves a local evidence extraction design. Compared with directly inputting full papers, smaller text windows reduce noise from cited work, irrelevant appendix tables and descriptions of prior methods. They also improve request stability and help control extraction cost. The 300-character overlap reduces the risk that a GPU model, count and usage context are split across different chunks. This helps ensure that key evidence is fully retained in at least one window. After extraction, we merged duplicate evidence at the paper-level and retained evidence spans traceable to the original text. These spans support later manual inspection and error analysis.

\subsection{GPU Model Normalization}
\label{app:Standardization}
\textbf{Hardware Catalog Mapping.} We mapped each extracted raw hardware name, recorded as \texttt{raw\_hardware\_name}, to a unique benchmark entry in the standard hardware catalog, recorded as \texttt{benchmark\_gpu\_name}. The catalog uses \texttt{Hardware name} as the primary key and includes fields for manufacturer, product family, generation, memory, bandwidth, and peak performance. We used a conservative matching principle. A raw name was assigned to a catalog entry only when it could be resolved to a specific GPU, TPU, or NPU device model. Generic hardware terms, model names, framework names, cloud instance types, and descriptions that only reported memory capacity were not directly mapped to specific hardware.

\textbf{Name Cleaning and Alias Matching.} We first normalized the raw names by standardizing case, cleaning manufacturer and model formats, normalizing memory notation, and removing non-discriminative suffixes such as ``GPU'', ``accelerator'', and ``device''. The cleaned names were then matched to the catalog in two ways. First, we performed exact matching against the normalized form of \texttt{Hardware name}. Second, we used an alias table derived from the catalog. If an alias mapped to only one catalog entry, the match was accepted. If the same alias mapped to multiple candidate devices, the record was marked as ambiguous and was not forced into a single normalized entry.

\textbf{Rule-Based Resolution of Common Hardware Mentions.} For common hardware names with unstable surface forms, we added a rule layer. Family-level or shorthand mentions such as A100, H100, H800, V100, P100, RTX 3090, RTX A6000, A800, and MI250X were resolved using model strings, manufacturer terms, memory information, and available catalog entries. When memory information distinguished variants, memory constraints were used first. For example, V100 could be mapped to the 16 GB or 32 GB variant, and A100 could be mapped to the 40 GB or 80 GB variant. If the text reported only the model family and did not provide enough variant information, we applied project-defined default variant rules. These cases were marked in \texttt{normalize\_reason} as \texttt{manual\_default\_variant\_rule}.

\textbf{Manual Catalog Supplementation.} When a confirmed hardware model or variant was missing from the catalog, we did not discard the record. Instead, we manually checked and supplemented the catalog using vendor naming conventions, product family relationships, memory information, and hardware generation information. Manual additions followed two principles. First, a catalog entry was added or corrected only when the model, manufacturer, and product category could be uniquely determined. Second, each added entry had to include the key fields needed for downstream analysis, including \texttt{Hardware name}, \texttt{Manufacturer}, \texttt{Generation}, \texttt{Family}, memory fields, and available performance fields. Supplemented entries were then incorporated into the same catalog and used in later normalization through the same exact matching, alias matching, and rule-based matching procedure.

\textbf{Unresolved Records and Exclusions.} Records that could not be reliably mapped were retained as unresolved, with the reason recorded. Typical unresolved cases included generic terms such as ``GPU'', ``TPU'', or ``NVIDIA GPU''; capacity-only descriptions such as ``80GB GPU''; cloud instance names rather than hardware model names; and extraction outputs that were actually model or framework names, such as BERT, LLaMA, Qwen, or PyTorch. CPU-related records were identified and removed during the renormalization stage and were not included in GPU catalog analyses.

\textbf{Traceability of Normalized Hardware Entries.} Thus, \texttt{benchmark\_gpu\_name} was not determined directly by the language model. It was determined by the hardware catalog, alias table, special model rules, memory variant rules, default variant configuration, and manually supplemented catalog entries. Each record retains \texttt{gpu\_name}, \texttt{benchmark\_gpu\_name}, \texttt{normalize\_status}, and \texttt{normalize\_reason}, allowing the mapping from original text to standard hardware entry to be traced.

\begin{table*}[t]
\centering
\footnotesize
\setlength{\tabcolsep}{4pt}
\begin{tabular}{@{}p{0.30\textwidth}rrrrrrrr@{}}
\toprule
\makecell[l]{\bfseries Reporting outcome} 
& {\bfseries\boldmath $N$} 
& \makecell[c]{\bfseries Event\\rate} 
& \makecell[c]{\bfseries\boldmath LPM\\$R^2$} 
& \makecell[c]{\bfseries\boldmath $\Delta$ LPM\\$R^2$} 
& \makecell[c]{\bfseries Logit\\AUC} 
& \makecell[c]{\bfseries\boldmath $\Delta$\\AUC} 
& \makecell[c]{\bfseries\boldmath McFadden\\pseudo-$R^2$} 
& \makecell[c]{\bfseries\boldmath $\Delta$ pseudo-\\$R^2$} \\
\midrule
Reports at least one standardizable GPU model 
& 13,745 & 0.5011 & 0.0804 & 0.0009 & 0.6572 & 0.0014 & 0.0602 & 0.0007 \\

Reports standardizable GPU model and quantity 
& 13,745 & 0.3875 & 0.0910 & 0.0013 & 0.6768 & 0.0015 & 0.0735 & 0.0011 \\
\bottomrule
\end{tabular}
\caption{Incremental explanatory power of organizational variables for GPU reporting sample membership.}
\label{tab:reporting-incremental-power}
\vspace{0.5em}
\begin{minipage}{0.95\textwidth}
\footnotesize
\textit{Note.} $\Delta$ statistics report the incremental explanatory power of organizational variables beyond the baseline specification. LPM denotes linear probability model. AUC denotes area under the ROC curve.
\end{minipage}
\end{table*}

\begin{table*}[t]
\centering
\footnotesize
\setlength{\tabcolsep}{6pt}

\begin{tabularx}{\textwidth}{@{}Xrrr@{}}
\toprule
\textbf{Predictor} 
& \makecell[c]{\bfseries Coef.\\(pp)} 
& \textbf{95\% CI} 
& {\bfseries\boldmath $p$-value} \\
\midrule

\multicolumn{4}{@{}l}{\textit{Outcome: Reports at least one standardizable GPU model}} \\
Industry participation 
& -0.935 & [-4.523, 2.654] & 0.610 \\
Industry--academia collaboration 
& 1.938 & [-2.287, 6.162] & 0.369 \\
Cross-sector collaboration 
& 2.329 & [-0.298, 4.957] & 0.082 \\
International collaboration 
& -1.595 & [-3.627, 0.436] & 0.124 \\
log(1 + number of organizations) 
& -2.320 & [-5.672, 1.031] & 0.175 \\

\midrule

\multicolumn{4}{@{}l}{\textit{Outcome: Reports standardizable GPU model and quantity}} \\
Industry participation 
& 1.012 & [-2.480, 4.504] & 0.570 \\
Industry--academia collaboration 
& 1.113 & [-3.016, 5.243] & 0.597 \\
Cross-sector collaboration 
& 2.354 & [-0.199, 4.907] & 0.071 \\
International collaboration 
& -1.332 & [-3.301, 0.637] & 0.185 \\
log(1 + number of organizations) 
& -2.726 & [-5.977, 0.524] & 0.100 \\

\bottomrule
\end{tabularx}
\caption{Full linear probability model coefficients for GPU reporting sample membership.}
\label{tab:reporting-lpm-coefficients}
\vspace{0.5em}
\begin{minipage}{0.95\textwidth}
\footnotesize
\textit{Note.} Coefficients are reported in percentage points. Models control for year, venue, and topic. Confidence intervals are 95\% intervals.
\end{minipage}
\end{table*}

\section{Reporting Coverage and Measurement Scope}

\subsection{Reporting Missingness and Organizational Predictors}
\label{app:reporting-missingness}

\textbf{Sample Construction after TPU Exclusion.} Because analyses of standardized GPU capacity depend on whether papers report standardizable GPU information, we further tested whether sample membership was systematically associated with organizational structure and collaboration characteristics. We first excluded papers with TPU matches in any row-level hardware name, benchmark name, generation or family field. This removed 167 papers containing TPU records. After excluding TPU papers and merging organization variables and topic labels, the final analysis sample contained 13,745 papers.

\textbf{Reporting Outcomes.} We constructed two binary outcome variables. The first indicates whether a paper reports at least one standardizable GPU model, which defines the model-reported sample. The second indicates whether a paper reports both a standardizable GPU model and an explicit GPU count, which defines the strict sample. In the final sample, 6,888 papers belonged to the model-reported sample and 6,857 did not, giving a positive class share of 50.11\%. For the strict sample, 5,326 papers belonged to the sample and 8,419 did not, giving a positive class share of 38.75\%.

\textbf{Model Specification and Fit Measures.} We estimated both linear probability models and logit models at the paper-level. The baseline model controlled for publication year, venue and topic fixed effects. The full model further added organizational and collaboration variables, including industry participation, collaboration between industry and academia, collaboration across sectors, international collaboration and the log transformed number of institutions. By comparing the baseline and full models, we assessed whether organizational variables explained the probability that a paper entered the model-reported sample or the strict sample. Model fit was evaluated using \(R^2\) for the linear probability model, AUC for the logit model and McFadden’s pseudo \(R^2\).

\textbf{Organizational Predictors of Reporting.} \cref{tab:reporting-incremental-power} and \cref{tab:reporting-lpm-coefficients} report the core organizational coefficients from the full linear probability models. Coefficients are expressed in percentage points. After controlling for year, venue and topic, industry participation, collaboration between industry and academia, collaboration across sectors, international collaboration and the number of institutions did not reach conventional levels of statistical significance. For the model-reported sample, the coefficient for collaboration across sectors was 2.329 percentage points, but the confidence interval crossed zero and the \(p\) value was 0.082. For the strict sample, the coefficient for collaboration across sectors was 2.354 percentage points, with a \(p\) value of 0.071. These estimates should therefore be interpreted only as weak signals, not as evidence for a strong association. The coefficients for the number of institutions were negative in both models, but they were unstable and not statistically significant.

\textbf{Implications for Conditional Interpretation.} Overall, this test supports a conditional interpretation of the main analyses. The standardized GPU capacity analyses do not claim to cover the true computational resource use of all NLP papers. They describe papers that report observable, extractable and standardizable GPU information. At the same time, after excluding TPU papers and controlling for year, venue and topic, industry participation, collaboration between industry and academia, collaboration across sectors, international collaboration and institution count add only limited predictive information for sample membership. The main associations between organizational background and reported computational resources are therefore unlikely to be explained solely by disclosure differences along these organizational dimensions. However, because GPU disclosure remains incomplete, all conclusions about computational capacity should still be understood as conditional findings based on the reporting sample.

\subsection{Audit of Compute-Consumption Reporting}
\label{app:compute_audit}
To assess whether reported GPU information could be extended into a measure of actual compute consumption, we conducted a manual audit of 240 papers that reported GPU usage. We sampled 15 papers from each of 16 venue--year strata to reduce the possibility that the audit was driven by reporting practices specific to a particular venue or year. For each sampled paper, we inspected the extracted GPU-related evidence passages and recorded whether they contained any information that could potentially inform computational consumption, such as execution duration, numbers of runs, or token usage.

As shown in \cref{tab:consumption_audit}, such a signal was visible in 92 of the 240 papers (38.3\%), whereas no consumption-related signal was visible in 148 papers (61.7\%). Importantly, the available information was not reported in a standardized form. Different papers reported different subsets of duration, run counts, token usage, or related quantities, and these signals generally could not be combined into a common quantity such as GPU-hours without introducing additional assumptions.

\begin{table}[t]
\centering
\small
\begin{tabular}{lrr}
\toprule
Audit outcome & Papers & Share \\
\midrule
Consumption-related signal visible & 92 & 38.3\% \\
No consumption-related signal visible & 148 & 61.7\% \\
\midrule
Total & 240 & 100.0\% \\
\bottomrule
\end{tabular}
\caption{Manual audit of consumption-related reporting among 240 papers reporting GPU usage. }
\label{tab:consumption_audit}
\end{table}

This audit indicates that constructing a corpus-wide measure of realized compute consumption would require restricting the analysis to a substantially smaller and selectively reported subset of papers. Such restriction could introduce considerable, and plausibly non-random, missingness because detailed consumption information is unlikely to be reported uniformly across research settings. We therefore do not infer actual compute consumption from these incomplete signals. Instead, the main analyses use GPU model and count to characterize the largest reported GPU hardware configuration, which we interpret as a measure of reported GPU capacity rather than realized computational consumption.

\section{Supplementary Results on Reported GPU Resources}
\label{app:supplementary-results}

\subsection{GPU Reporting Completeness}
\label{app:reporting-completeness}

\begin{figure*}[t]
  \centering
  \includegraphics[width=0.95\textwidth]{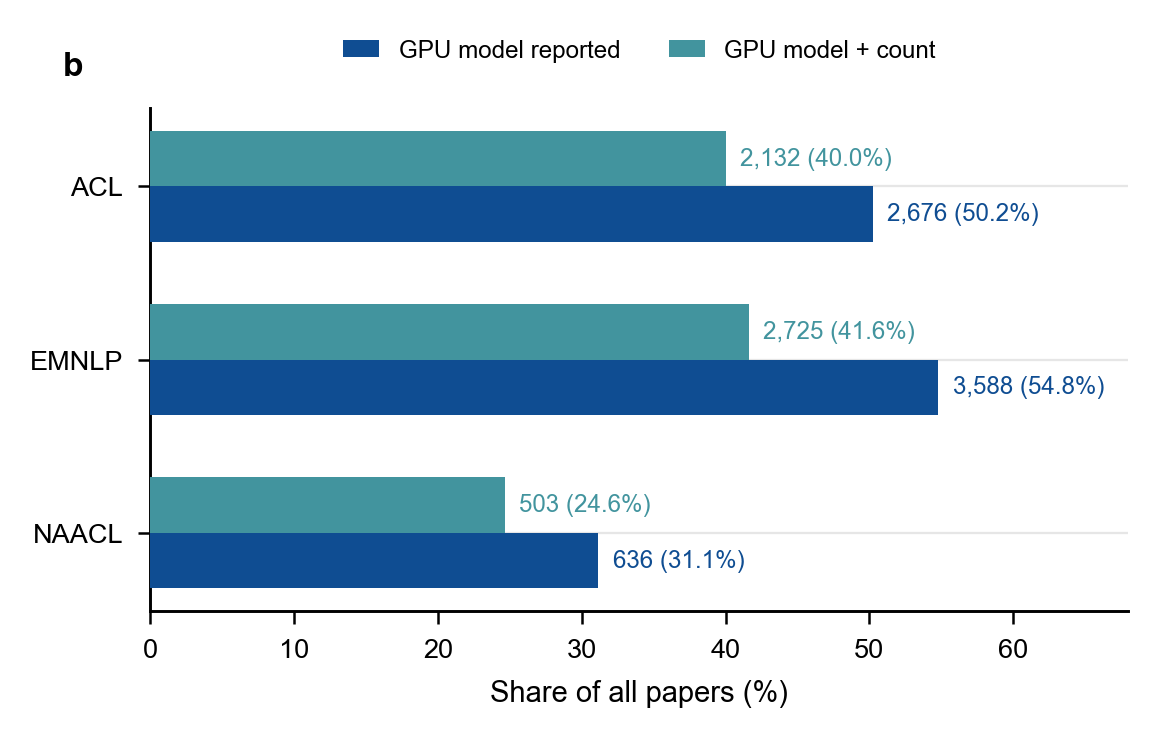}
  \caption{GPU hardware reporting completeness across venues.}
  \label{fig:reporting-time}
\end{figure*}

\cref{fig:reporting-time} shows venue-level differences. EMNLP and ACL have higher GPU model reporting rates, at 54.8\% and 50.2\%, respectively, and similar rates of joint model-and-count reporting, at 41.6\% and 40.0\%. NAACL has lower rates on both measures, at 31.1\% for GPU model reporting and 24.6\% for joint reporting. Overall, GPU resources have become more visible in NLP papers, but reporting remains incomplete and varies across venues.

\subsection{Reported GPU Counts and Hardware Generations}
\label{app:gpu-counts-generations}

\begin{figure*}[t]
  \centering
  \includegraphics[width=0.95\textwidth]{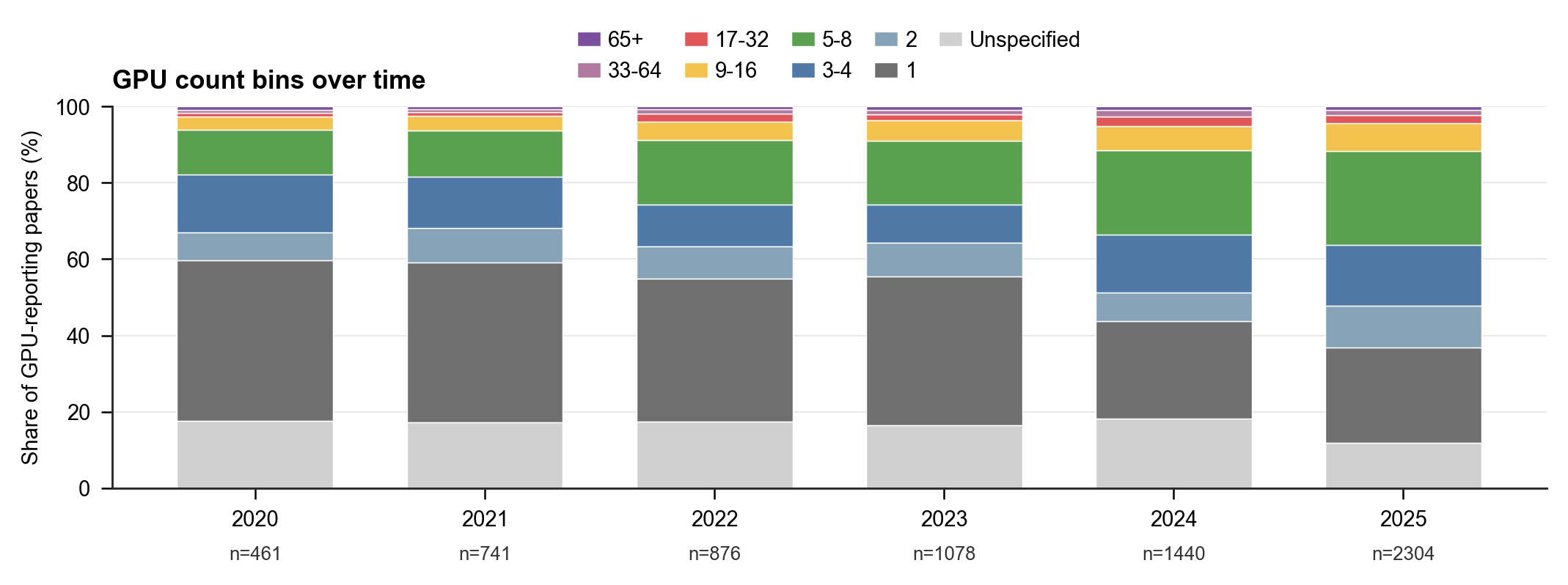}
  \caption{Reported GPU count distributions over time.}
  \label{fig:gpu-count}
\end{figure*}

\cref{fig:gpu-count} shows the annual distribution of reported GPU count bins among papers that reported GPU resources. The share of single-GPU papers declined over time, while multi-GPU configurations became more common. Medium-scale configurations, especially 3--4 GPUs, 5--8 GPUs, and 9--16 GPUs, increased noticeably. This pattern indicates that reported hardware scale has grown, but not through universal hyperscaling: papers reporting 33--64 GPUs or 65+ GPUs remained rare throughout the period.

\begin{figure*}[t]
  \centering
  \includegraphics[width=0.95\textwidth]{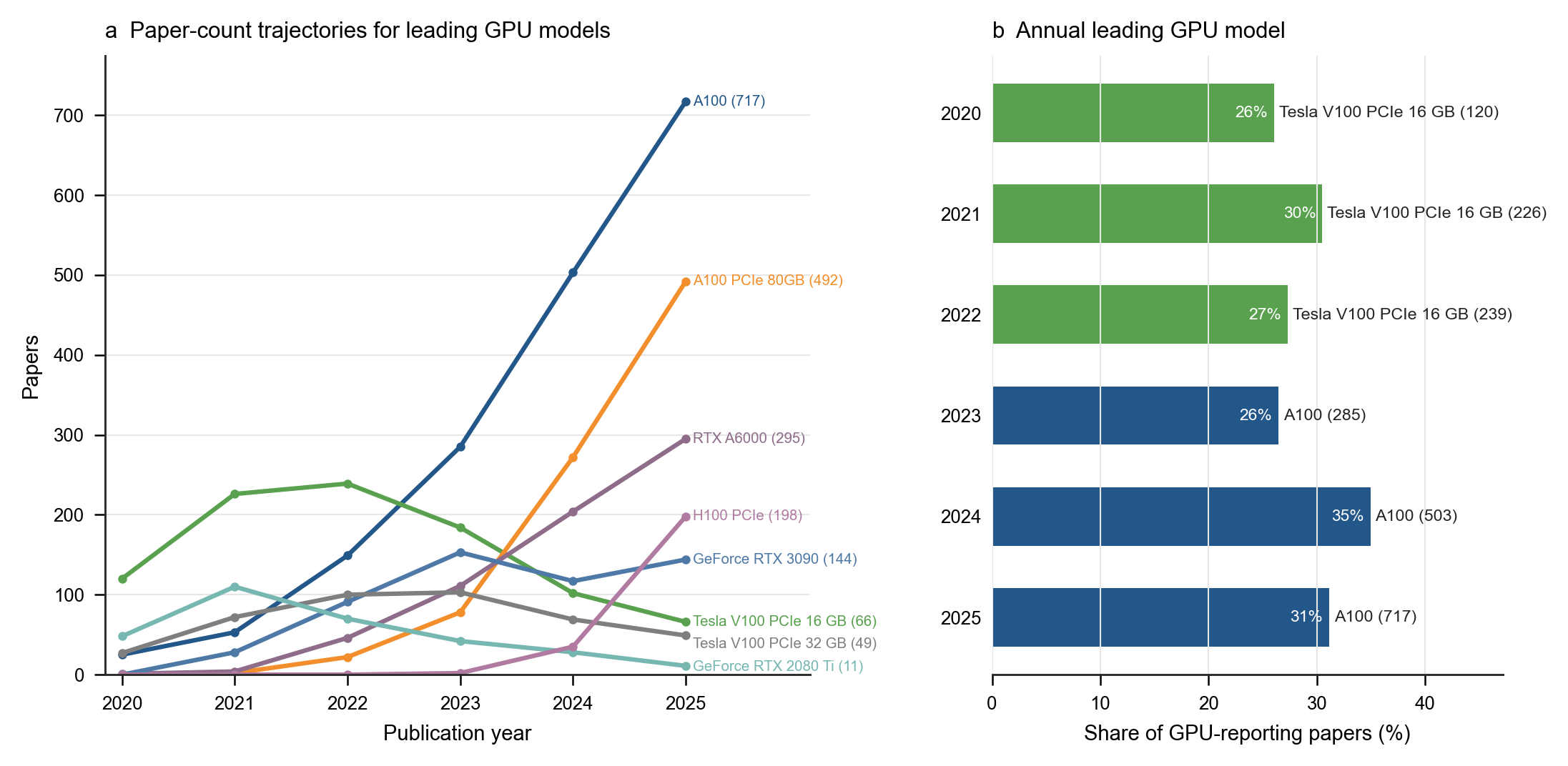}
  \caption{Transition of leading reported GPU models in NLP papers.}
  \label{fig:top-model}
\end{figure*}

\cref{fig:top-model} reports changes in the most frequently reported GPU models. \hyperref[fig:top-model]{\cref*{fig:top-model} (a)} shows a clear generational transition. Tesla V100 PCIe 16 GB was the dominant model in the early years, peaking in 2021 and 2022 before declining. A100 GPUs increased rapidly after 2022 and became the most frequently reported model by 2025. Newer and higher-performance models, including A100 PCIe 80 GB, RTX A6000, and H100 PCIe, also became more visible in later years.

\hyperref[fig:top-model]{\cref*{fig:top-model} (b)} identifies the leading GPU model in each year. From 2020 to 2022, Tesla V100 PCIe 16 GB was the most common reported GPU. From 2023 onward, A100 replaced V100 as the leading model and remained the most common GPU in 2024 and 2025.

\subsection{Reported GPU Memory Capacity}
\label{app:gpu-memory}

\begin{figure*}[t]
  \centering
  \includegraphics[width=0.95\textwidth]{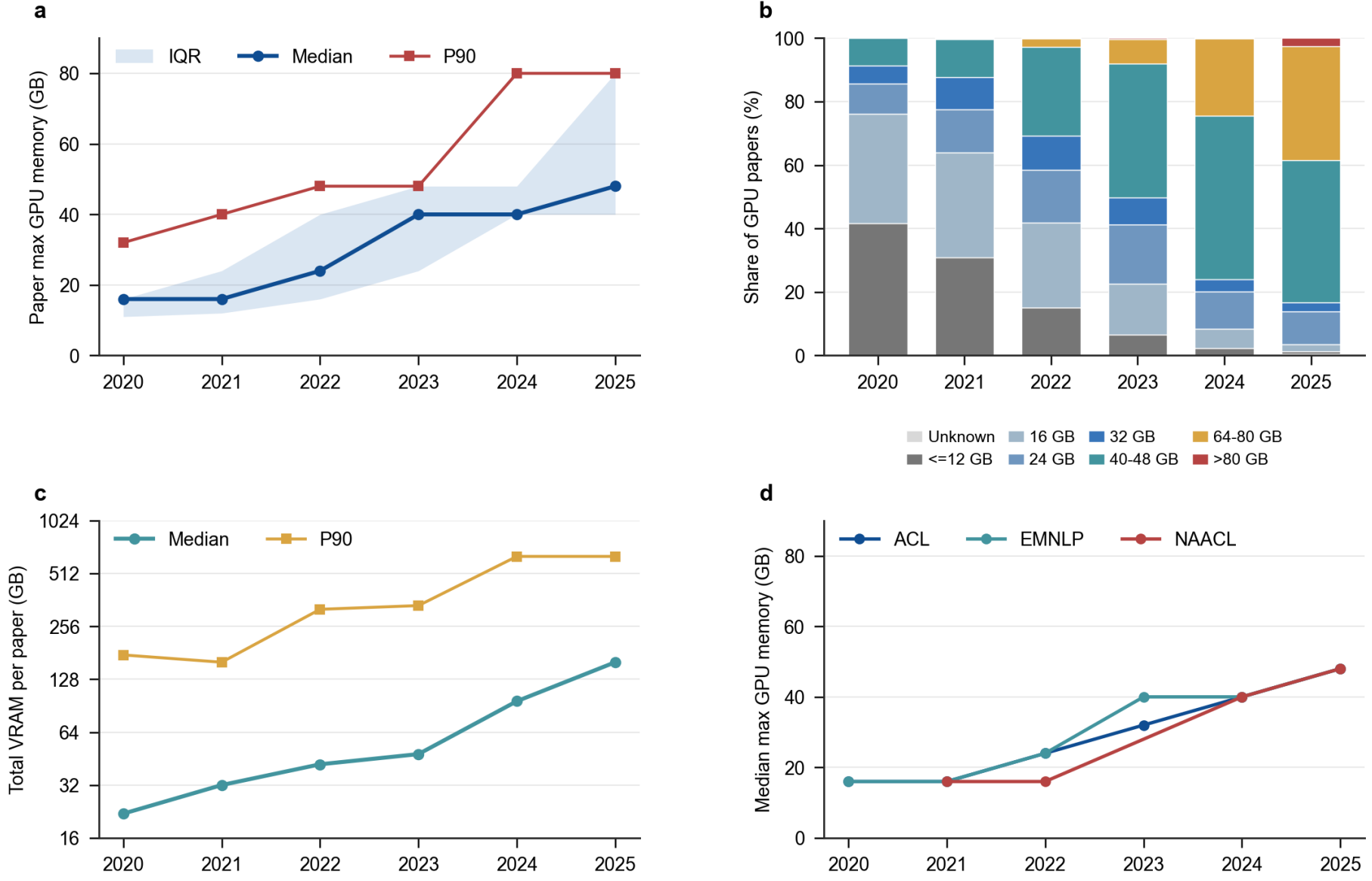}
  \caption{Growth in reported GPU memory capacity in NLP papers.}
  \label{fig:memory}
\end{figure*}

\cref{fig:memory} reports changes in GPU memory configurations from 2020 to 2025. \hyperref[fig:memory]{\cref*{fig:memory} (a)} shows that the maximum GPU memory reported at the paper-level increased steadily. The median rose from about 16 GB in 2020--2021 to nearly 48 GB in 2025, while the P90 increased from about 32 GB to 80 GB. This indicates that high-memory GPUs became increasingly common in the reporting sample.

\hyperref[fig:memory]{\cref*{fig:memory} (b)} shows a corresponding shift in memory categories. Early papers had higher shares of unknown memory, low-memory GPUs, and 16 GB or 32 GB configurations. After 2023, the shares of 40--48 GB and 64--80 GB configurations increased markedly, reflecting a transition from the V100 period toward higher-memory devices such as A100 and H100 GPUs. Panel (c) shows a similar trend for total GPU memory at the paper-level. As both GPU counts and per-GPU memory increased, the median and P90 of total VRAM also rose substantially. Panel (d) shows that the median maximum GPU memory increased across conferences, suggesting that this trend was not limited to a single venue.

\subsection{Distribution of Reported GPU Capacity}
\label{app:gpu-capacity-distribution}

\begin{figure}[t]
  \centering
  \includegraphics[width=\columnwidth]{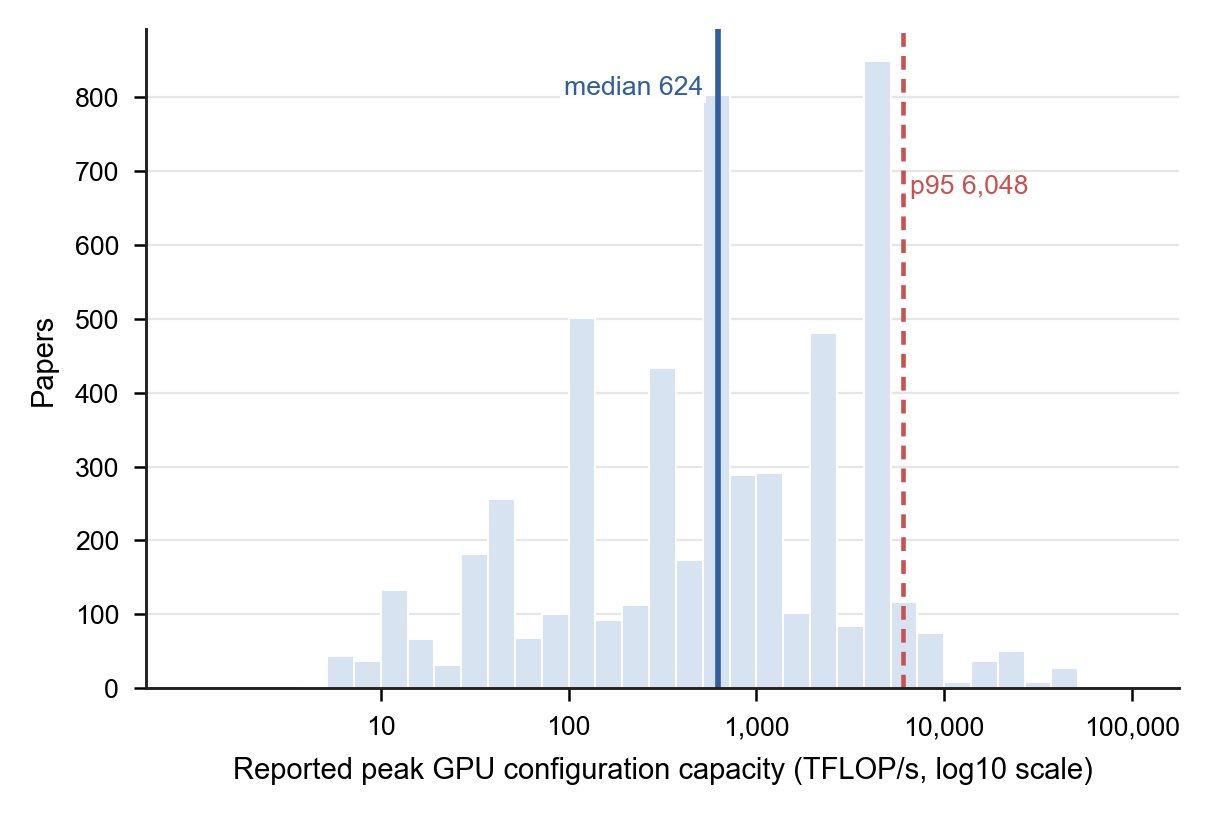}
  \caption{Distribution of reported peak GPU configuration capacity in NLP papers.}
  \label{fig:gpu-capacity}
\end{figure}

\cref{fig:gpu-capacity} shows the distribution of reported peak GPU configuration capacity. The x-axis reports the maximum GPU configuration capacity at the paper-level, measured in TFLOP/s and plotted on a log$_{10}$ scale. Reported GPU capacity is strongly right-skewed. Most papers fall in a medium-capacity range, with a median of 624 TFLOP/s, indicating that the typical reported configuration is not a very large GPU cluster. At the same time, the right tail is substantial: the P95 reaches 6,048 TFLOP/s, showing that a small number of papers report configurations far above the median.

\begin{figure*}[t]
  \centering
  \includegraphics[width=0.95\textwidth]{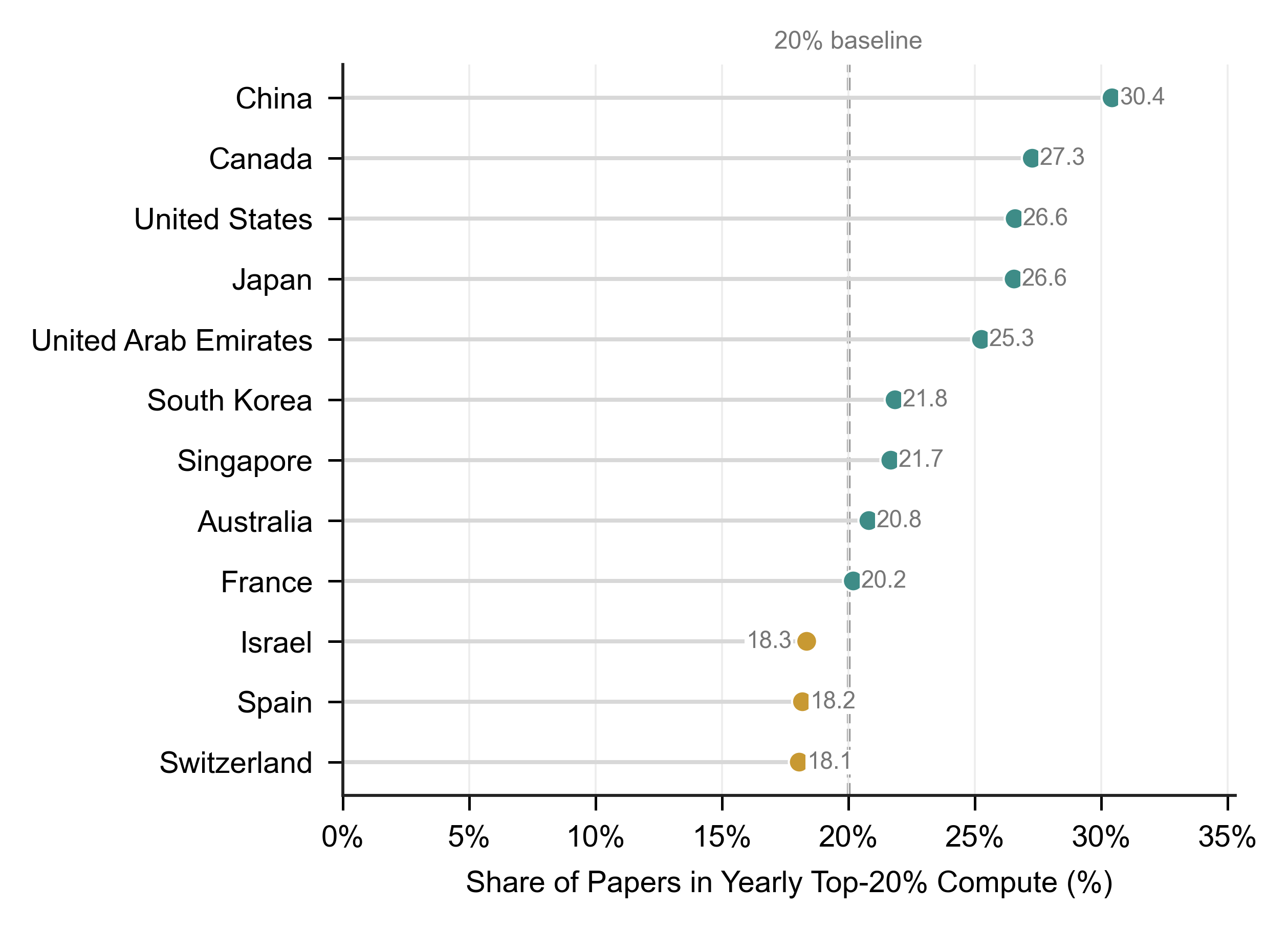}
  \caption{Country-level concentration in high reported GPU compute.}
  \label{fig:gpu-country}
\end{figure*}

\begin{figure*}[t]
  \centering
  \includegraphics[width=0.95\textwidth]{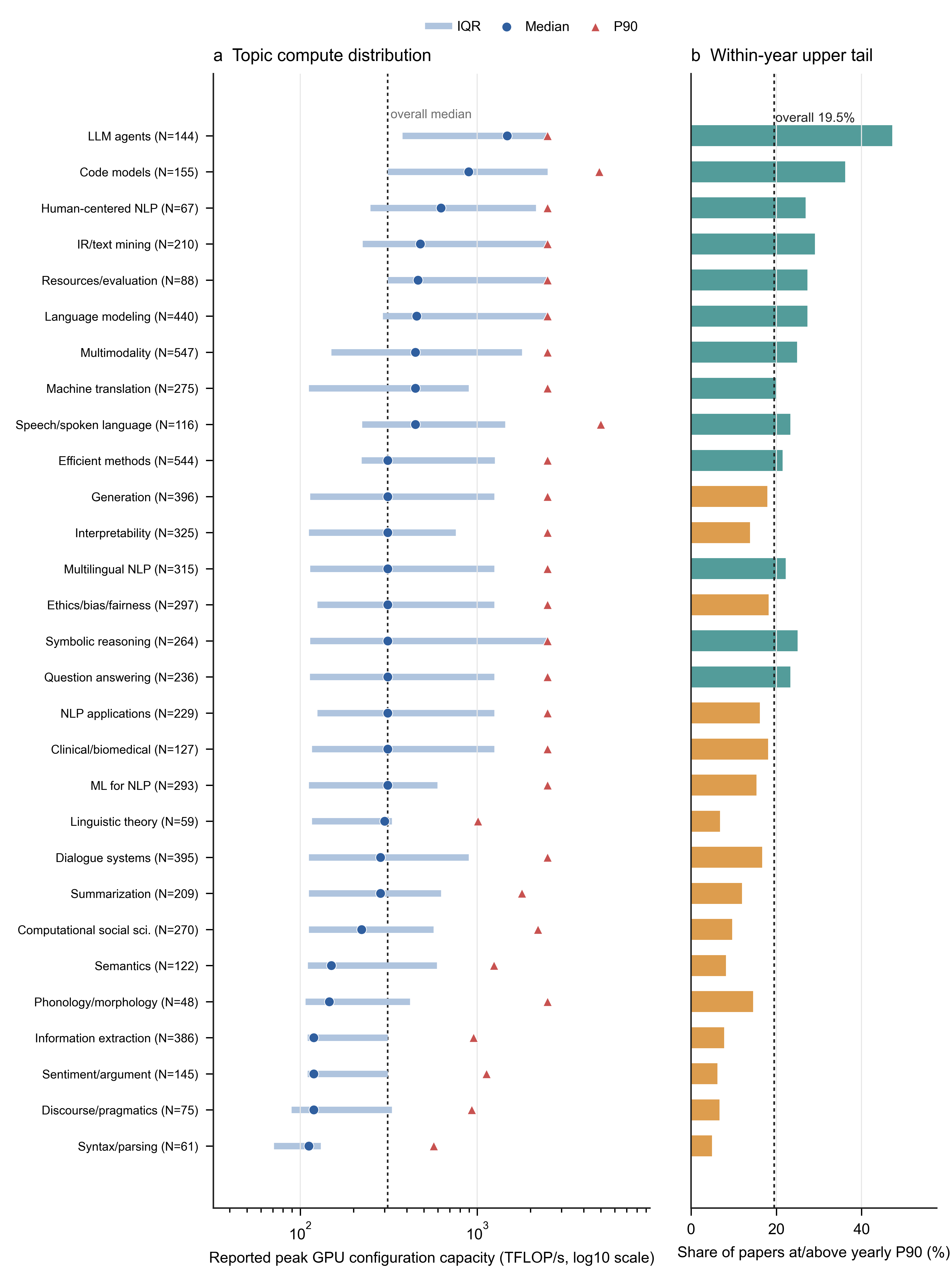}
  \caption{Reported GPU hardware capacity across NLP research topics. Panel (a) reports the median, interquartile range, and P90 of paper-level peak reported GPU configuration capacity. Panel (b) reports the share of papers at or above the year-specific P90 capacity threshold. Because ties at the P90 threshold are retained, the overall share can exceed 10\%; the dashed line indicates the empirical overall share of 19.5\%.}
  \label{fig:topic}
\end{figure*}

\subsection{Institutional Correlates of Reported GPU Capacity}
\label{app:regression-institution}

\textbf{Regression Sample and Outcome Variable.} We use the strict sample for the regression analysis. This sample includes papers that report both a standardizable GPU model and an explicit GPU count, yielding 5,360 papers. After removing papers that could not be matched to the relevant organization variables, the final regression sample contains 5,357 papers. The dependent variable is the maximum observable GPU configuration capacity at the paper-level, defined as follows:
\begin{equation}
\label{eq:y}
Y_i = \log_{10}\left({\texttt{GPU\_Capacity}_i}\right)
\end{equation}
Here, \(\mathit{GPU\_Capacity}_i\) is defined in \cref{eq:gpu-capacity}. We take the base 10 logarithm of this measure to reduce distributional skewness.

\textbf{Institutional Predictors.} The regression models examine five organizational characteristics: industry participation, collaboration between industry and academia, collaboration across sectors, international collaboration, and the number of participating organizations. The first four variables are binary indicators. The number of participating organizations is measured as \(\log(1+\mathrm{Organizations})\).

\textbf{Fixed Effects and Conditional Interpretation.} All models control for year, topic and venue fixed effects. Year fixed effects account for changes in overall GPU capacity across publication years. Topic fixed effects account for differences in compute demand across NLP research areas. Venue fixed effects account for systematic differences across conferences. The coefficients therefore estimate the association between organizational structure and reported peak GPU configuration capacity, conditional on year, topic and venue differences.

\textbf{Single-Predictor Models.} Models M1 to M5 are single predictor models. Each model includes one organizational characteristic and controls for year, topic and venue fixed effects:
\begin{equation}
\label{eq:y-1}
Y_i = \alpha + \beta X_i + \gamma_{\mathrm{year}} + \delta_{\mathrm{topic}} + \lambda_{\mathrm{venue}} + \varepsilon_i
\end{equation}
where \(X_i\) denotes one of the five organizational characteristics.

\textbf{Full Conditional Model.} Model M6 is the full conditional model and includes all five organizational characteristics:
\begin{equation}
\label{eq:y-2}
\begin{aligned}
Y_i ={}&
\alpha
+ \beta_1 \mathrm{Industry}_i \\
&+ \beta_2 \mathrm{IndustryAcademia}_i
+ \beta_3 \mathrm{CrossSector}_i \\
&+ \beta_4 \mathrm{International}_i\\
&+ \beta_5 \log(1+\mathrm{Organizations}_i) \\
&+ \gamma_{\mathrm{year}}
+ \delta_{\mathrm{topic}}
+ \lambda_{\mathrm{venue}}
+ \varepsilon_i.
\end{aligned}
\end{equation}

\textbf{Effect Size Conversion.} In the full model, each coefficient estimates the conditional association between that organizational characteristic and reported peak GPU configuration capacity, after controlling for the other organizational characteristics and for year, topic, and venue fixed effects. Because the dependent variable is in \(\log_{10}\) form, a coefficient \(\beta\) can be converted into a percentage difference in reported GPU capacity as:
\begin{equation}
\label{eq:percentage-conversion}
(10^{\beta} - 1) \times 100\%.
\end{equation}

\textbf{Regression Results.} Regression results (\cref{tab:institutional_compute_regression}) are consistent with this pattern, while also showing that institutional indicators partly overlap with one another. In models that include one focal institutional variable at a time and control for year, venue, and topic, industry participation is associated with 84.8\% higher reported GPU capacity. Industry--academia and cross-sector collaborations are associated with increases of 56.8\% and 44.6\%, respectively. The number of participating institutions is also positively associated with reported capacity, whereas international collaboration alone is small and not statistically significant. When all institutional variables are included simultaneously, industry participation remains strongly positive, while the coefficient for industry--academia becomes negative, suggesting that its positive bivariate association is partly absorbed by industry participation and other collaboration structures.

\begin{table*}[t]
\centering
\small
\setlength{\tabcolsep}{5pt}
\renewcommand{\arraystretch}{1.08}
\begin{tabular}{lcccccc}
\toprule
\textbf{Predictor} 
& \textbf{M1} & \textbf{M2} & \textbf{M3} 
& \textbf{M4} & \textbf{M5} & \textbf{M6} \\
\midrule
Industry 
& $0.267^{***}$ &  &  &  &  & $0.474^{***}$ \\
& $(0.016)$ &  &  &  &  & $(0.042)$ \\

Industry-academia 
&  & $0.195^{***}$ &  &  &  & $-0.299^{***}$ \\
&  & $(0.017)$ &  &  &  & $(0.047)$ \\

Cross-sector 
&  &  & $0.160^{***}$ &  &  & $0.057^{*}$ \\
&  &  & $(0.015)$ &  &  & $(0.023)$ \\

International collaboration 
&  &  &  & $0.030$ &  & $-0.023$ \\
&  &  &  & $(0.016)$ &  & $(0.019)$ \\

$\log(1+\text{organization count})$ 
&  &  &  &  & $0.149^{***}$ & $0.085^{**}$ \\
&  &  &  &  & $(0.022)$ & $(0.031)$ \\

\midrule
Year FE  & Yes & Yes & Yes & Yes & Yes & Yes \\
Topic FE & Yes & Yes & Yes & Yes & Yes & Yes \\
Venue FE & Yes & Yes & Yes & Yes & Yes & Yes \\
Other institutional controls & No & No & No & No & No & Yes \\
\midrule
Observations & 5,357 & 5,357 & 5,357 & 5,357 & 5,357 & 5,357 \\
$R^2$ & 0.191 & 0.170 & 0.164 & 0.148 & 0.155 & 0.201 \\
Adjusted $R^2$ & 0.186 & 0.165 & 0.159 & 0.142 & 0.150 & 0.195 \\
Implied \% change & +84.8\% & +56.8\% & +44.6\% & +7.0\% & +41.0\% & See note \\
\bottomrule
\end{tabular}

\caption{Associations between institutional characteristics and reported GPU capacity.}
\label{tab:institutional_compute_regression}

\vspace{0.5em}
\begin{minipage}{0.98\textwidth}
\footnotesize
\textit{Note.} 
Models M1--M5 estimate one focal institutional variable at a time. Model M6 includes all five institutional variables simultaneously. 
All models are estimated using OLS with year, topic, and venue fixed effects. HC3 robust standard errors are reported in parentheses. 
The implied percentage change is calculated as $100 \times (10^{\beta}-1)$. 
In M6, the implied changes are: Industry +197.8\%, Industry-academia $-49.8$\%, Cross-sector +13.9\%, International collaboration $-5.2$\%, and $\log(1+\text{organization count})$ +21.6\%. 
$^{*}p<0.05$, $^{**}p<0.01$, $^{***}p<0.001$.
\end{minipage}

\end{table*}

\subsection{Regional Correlates of Reported GPU Capacity}
\label{app:instifutional-correlate}
\cref{fig:gpu-country} reports the share of papers from each country or region that fall into the top 20\% of reported GPU capacity within their publication year. The dashed line marks the 20\% benchmark. If papers from a country or region were distributed in the high-compute tail in the same way as the overall sample, its share would be close to this line.

Papers involving China, Canada, the United States, Japan, and the United Arab Emirates are more likely to enter the annual top 20\% of reported GPU capacity. China has the highest share, at 30.4\%. South Korea, Singapore, Australia, and France are also slightly above the 20\% benchmark. By contrast, Israel, Spain, and Switzerland fall below 20\%, indicating lower representation in the high-compute tail.

\subsection{Reported GPU Capacity Across Research Contexts}
\label{app:gpu-capacity-contexts}

\cref{fig:topic} shows differences in reported GPU configuration capacity across NLP research topics. \hyperref[fig:topic]{\cref*{fig:topic} (a)} compares topics using paper-level peak GPU configuration capacity, reporting the median, interquartile range, and P90. Reported GPU capacity is well above the overall median in topics such as LLM agents, Code models, Human-centered NLP, IR/text mining, Resources/evaluation, Language modeling, and Multimodality. This indicates that high-capacity configurations are more concentrated in research on large models, code models, multimodal systems, and resource- or evaluation-oriented work. \hyperref[fig:topic]{\cref*{fig:topic} (b)} reports the share of papers in each topic whose reported GPU capacity is at or above the year-specific P90 threshold. Because reported GPU capacities are discrete and many papers are tied at the annual P90 threshold, this inclusive definition yields an overall share of 19.5\%, rather than exactly 10\%. The dashed line marks this empirical overall share. Topics such as LLM agents and Code models exceed this benchmark by a substantial margin. By contrast, Syntax/parsing, Discourse/pragmatics, Sentiment/argument mining, Information extraction, and Semantics have lower shares in the P90-threshold group.

\section{Supplementary Impact Analyses}
\label{app:regression-Impact}

\subsection{Impact Outcomes and Analysis Samples}
\label{app:impact-outcomes}

The citation analyses use the strict 2020--2023 sample to reduce
citation-window truncation. The strict sample requires both a
standardizable GPU model and an explicit GPU count. After excluding
papers with incomplete citation or control variables, the citation
regression sample contains 2,194 papers. Because award status is
observed at publication, the award analyses use the 
2020--2025 sample of 5,357 papers, including 111 award-positive papers.

Our primary impact outcome is the NLP topic--year citation percentile.
Each paper is ranked relative to papers published in the same year and assigned to the same primary NLP topic, with average ranks used for citation ties and the resulting ranks scaled to the unit interval. For paper \(i\) published in year \(y\) and assigned to topic \(t\),
let \(r_i\) denote its ascending citation rank within the corresponding
topic--year reference cell, with average ranks used for ties, and let
\(n_{ty}\) denote the number of papers in that cell. We define
\begin{equation}
\label{eq:topic-year}
P_i^{\mathrm{NLP}}
=
\frac{r_i-0.5}{n_{ty}}
\end{equation}
Higher values therefore indicate greater citation impact relative to
papers from the same publication year and NLP topic. This reference set accounts for both citation-window differences across publication years and variation in citation practices across NLP research areas. The OpenAlex field-normalized citation percentile is used as a secondary normalized outcome based on an external reference classification. Complementary outcomes are
$\log(1+\mathrm{citations})$, raw citation counts, an indicator for
belonging to the citation top 10\% within publication-year-by-venue
cells, and paper-award status.

All baseline models include publication-year-by-venue fixed effects,
primary-topic fixed effects, team-size controls, and organization-count
controls. HC3 robust standard errors are used for OLS and linear
probability models, and HC0 robust standard errors are used for PPML
models. The estimates are conditional associations and are not
interpreted as causal effects of computational resources.

\subsection{Sensitivity to Alternative High-Capability and High-Impact Cutoffs}
\label{app:sensitivity-cutoff}

The binary comparison in the main analysis requires operational thresholds for both reported GPU capability and citation impact. Our baseline specification defines high reported GPU capability as the yearly top 20\% among papers with positive, quantifiable reported GPU capability and high citation impact as the top 10\% of citations within each publication-year-by-venue group. GPU capability is ranked within publication year because the analysis concerns the annual distribution of reported hardware capability, whereas citation impact is ranked within publication-year-by-venue groups to account for differences in citation accumulation across publication cohorts and venues.

To assess sensitivity to these choices, let \(H^{\mathrm{GPU}}_a\) indicate membership in the top \(a\%\) of reported GPU capability within publication year, where \(a \in \{10,20,30\}\), and let \(H^{\mathrm{impact}}_b\) indicate membership in the top \(b\%\) of citations within publication-year-by-venue groups, where \(b \in \{5,10,20\}\). For each combination, we calculate the rate ratio

\begin{equation}
\label{eq:R}
R_{a,b} =
\frac{
\Pr\left(
H^{\mathrm{impact}}_b = 1
\mid
H^{\mathrm{GPU}}_a = 1
\right)
}{
\Pr\left(
H^{\mathrm{impact}}_b = 1
\mid
H^{\mathrm{GPU}}_a = 0
\right)
}.
\end{equation}

Thus, \(R_{a,b} > 1\) indicates that papers above the corresponding GPU-capability threshold have a higher rate of belonging to the high-impact group than the remaining GPU-quantifiable papers. Under the baseline specification (\(a=20\), \(b=10\)), 14.5\% of high-capability papers belong to the citation top 10\%, compared with 9.1\% of the remaining papers, yielding a rate ratio of 1.59. Across all nine threshold combinations reported in \cref{tab:cutoff-sensitivity}, the ratio ranges from 1.49 to 2.14 and remains above one.

We further test whether the result depends on dichotomizing reported GPU capability, as shown in \cref{tab:continuous-cutoff-sensitivity}. Specifically, we re-estimate the preferred linear probability model using the continuous \(\log_{10}\)-transformed maximum reported GPU capability while varying the high-impact definition across the citation top 5\%, 10\%, and 20\%. All specifications retain the same publication-year-by-venue fixed effects and topic, team-size, and organization-count controls as the baseline model. Because reported GPU capability is \(\log_{10}\)-transformed, each coefficient represents the percentage-point difference in the probability of belonging to the corresponding high-impact group associated with a tenfold increase in reported GPU capability.

\begin{table}[t]
\centering
\footnotesize
\setlength{\tabcolsep}{3pt}
\begin{tabular}{@{}lcccc@{}}
\toprule
Citation cutoff
& Coef. (pp)
& 95\% CI
& \(p\)
& \(\Delta R^2\) \\
\midrule
Top 5\%
& +4.04
& [1.81, 6.26]
& \(<0.001\)
& 0.0085 \\
Top 10\%
& +3.84
& [1.14, 6.53]
& 0.005
& 0.0043 \\
Top 20\%
& +7.20
& [3.80, 10.59]
& \(<0.001\)
& 0.0086 \\
\bottomrule
\end{tabular}
\caption{Sensitivity of the continuous reported GPU-capability model to alternative high-impact thresholds. Coefficients are percentage-point differences associated with a tenfold increase in reported GPU capability. All models include the same fixed effects and controls as the baseline specification.}
\label{tab:continuous-cutoff-sensitivity}
\end{table}

\begin{table*}[t]
\centering
\scriptsize
\setlength{\tabcolsep}{3.0pt}
\renewcommand{\arraystretch}{1.05}
\begin{tabular*}{\textwidth}{@{\extracolsep{\fill}}p{3.1cm}llrrrrlc@{}}
\toprule
Outcome & Estimator & $N$ & $\beta$ & SE & 95\% CI & $p$ & Implied effect & $\Delta R^2$ \\
\midrule
NLP topic--year percentile (primary)
& OLS & 2,194 & 0.0352 & 0.0115 & [0.0127, 0.0577] & 0.002
& +3.52 pp & 0.0042 \\
OpenAlex field-normalized percentile
& OLS & 2,194 & 0.0126 & 0.0088 & [$-0.0046$, 0.0298] & 0.151
& +1.26 pp & 0.0009 \\
$\log(1+\mathrm{citations})$
& OLS & 2,194 & 0.1719 & 0.0491 & [0.0757, 0.2681] & $<0.001$
& +18.8\% & 0.0052 \\
Citation count
& PPML & 2,194 & 0.4775 & 0.0852 & [0.3105, 0.6445] & $<0.001$
& +61.2\% & -- \\
Top-10\% cited
& LPM & 2,194 & 0.0384 & 0.0138 & [0.0114, 0.0654] & 0.005
& +3.84 pp & 0.0043 \\
Awarded
& LPM & 5,357 & 0.0086 & 0.0045 & [$-0.0002$, 0.0174] & 0.056
& +0.86 pp & 0.0011 \\
\bottomrule
\end{tabular*}
\caption{Full adjusted models relating aggregate reported GPU capability to scholarly impact. The implied effects correspond to a tenfold increase in aggregate reported GPU capability. For OLS log-citation and PPML models, percentage effects are calculated as $100\times(e^{\beta}-1)$. For percentile and linear-probability outcomes, effects are expressed in percentage points (pp). Incremental $R^2$ is calculated against a controls-only model estimated on the same sample; it is not reported for PPML.}
\label{tab:impact-aggregate-full}
\end{table*}

The continuous specifications remain positive and statistically significant under all three citation thresholds. We also re-estimated the corresponding binary high-capability models across all nine combinations of GPU-capability and citation thresholds, and the estimated association remained positive in every specification. Together, these results show that the positive association between reported GPU capability and high citation impact is robust to alternative threshold choices and is not an artifact of dichotomizing the GPU-capability measure.

\subsection{Aggregate and Dimension-Specific Associations}
\label{app:impact-full-models}

The aggregate specification uses the maximum reported paper-level GPU
capability:
\begin{equation}
\label{eq:compute-citation-aggregate}
\begin{aligned}
Y_i ={}& \alpha
+ \beta \log_{10}(\mathrm{GPUCapability}_i) \\
&+ \theta_{\mathrm{year}\times\mathrm{venue}}
+ \delta_{\mathrm{topic}}
+ \mathbf{X}_i^\top\gamma
+ \varepsilon_i .
\end{aligned}
\end{equation}
where \(\mathbf{X}_i\) contains the baseline paper-level controls for
team size and organization count, using the same transformations as in the main analysis.
A one-unit increase in the focal predictor corresponds to a tenfold
increase in aggregate reported GPU capability. The full estimates from this aggregate specification are reported in
\cref{tab:impact-aggregate-full}.

To distinguish deployment scale from hardware generation, we estimate a
separate joint model:
\begin{equation}
\label{eq:compute-citation-components}
\begin{aligned}
Y_i ={}& \alpha
+ \beta_N \log_{10}(\mathrm{GPUCount}_i) \\
&+ \beta_G \mathbf{1}\{\mathrm{AmpereOrNewer}_i\} \\
&+ \theta_{\mathrm{year}\times\mathrm{venue}}
+ \delta_{\mathrm{topic}}
+ \mathbf{X}_i^\top \gamma
+ \varepsilon_i .
\end{aligned}
\end{equation}
where \(\mathbf{X}_i\) contains the baseline paper-level controls for
team size and organization count, using the same transformations as in the main analysis.

Aggregate GPU capability is not included in this specification. The
GPU-count coefficient represents the association with a tenfold
increase in the reported count while holding hardware generation
constant. The generation coefficient compares Ampere-or-newer/equivalent hardware with earlier-generation hardware while holding GPU
count constant. The exact generation mapping is documented in
\cref{app:Standardization}. Full estimates from this joint component
specification are reported in \cref{tab:impact-components-full}.

\begin{table*}[t]
\centering
\scriptsize
\setlength{\tabcolsep}{3.0pt}
\renewcommand{\arraystretch}{1.05}
\begin{tabular*}{\textwidth}{@{\extracolsep{\fill}}p{3.1cm}lrrrrlc@{}}
\toprule
Outcome & Term & $\beta$ & SE & 95\% CI & $p$ & Implied effect & Joint $\Delta R^2$ \\
\midrule
NLP topic--year percentile
& GPU count & 0.0457 & 0.0133 & [0.0196, 0.0718] & $<0.001$ & +4.57 pp & 0.0093 \\
& Newer generation & 0.0416 & 0.0135 & [0.0151, 0.0681] & 0.002 & +4.16 pp &  \\
\addlinespace
OpenAlex field-normalized percentile
& GPU count & 0.0196 & 0.0101 & [$-0.0002$, 0.0394] & 0.052 & +1.96 pp & 0.0033 \\
& Newer generation & 0.0202 & 0.0105 & [$-0.0004$, 0.0408] & 0.055 & +2.02 pp &  \\
\addlinespace
$\log(1+\mathrm{citations})$
& GPU count & 0.2206 & 0.0571 & [0.1087, 0.3325] & $<0.001$ & +24.7\% & 0.0086 \\
& Newer generation & 0.1396 & 0.0552 & [0.0314, 0.2478] & 0.011 & +15.0\% &  \\
\addlinespace
Citation count, PPML
& GPU count & 0.5264 & 0.0845 & [0.3608, 0.6920] & $<0.001$ & +69.3\% & -- \\
& Newer generation & 0.2699 & 0.1321 & [0.0110, 0.5288] & 0.041 & +31.0\% &  \\
\addlinespace
Top-10\% cited
& GPU count & 0.0441 & 0.0162 & [0.0123, 0.0759] & 0.006 & +4.41 pp & 0.0051 \\
& Newer generation & 0.0217 & 0.0150 & [$-0.0077$, 0.0511] & 0.148 & +2.17 pp &  \\
\addlinespace
Awarded
& GPU count & 0.0133 & 0.0056 & [0.0023, 0.0243] & 0.017 & +1.33 pp & 0.0018 \\
& Newer generation & $-0.0025$ & 0.0053 & [$-0.0129$, 0.0079] & 0.645 & $-0.25$ pp &  \\
\bottomrule
\end{tabular*}
\caption{Full joint models separating reported GPU count from hardware generation. GPU count and the Ampere-or-newer/equivalent indicator are entered simultaneously. GPU-count effects correspond to a tenfold increase in the reported count; newer-generation effects compare Ampere-or-newer/equivalent hardware with earlier generations. Citation models use $N=2{,}194$ and the award model uses $N=5{,}357$. The joint incremental $R^2$ is the increase from adding both hardware terms to the controls-only model estimated on the same sample.}
\label{tab:impact-components-full}
\end{table*}

As shown in
\cref{tab:impact-aggregate-full,tab:impact-components-full}, the aggregate model is positively associated with the primary NLP
topic--year percentile, but adds only 0.0042 to model $R^2$. The
OpenAlex field-normalized estimate is smaller and statistically
imprecise. In the joint component model, both GPU count and newer
hardware generation are positively associated with the primary
within-NLP percentile. Across the complementary citation outcomes, GPU
count displays the more consistent positive pattern, whereas the
Ampere-or-newer/equivalent indicator is supported for citation intensity but not
for top-10\% citation status. Because the two hardware measures are on
different scales, their coefficient magnitudes are not interpreted as
a direct ranking of importance. The joint component model also adds
less than 0.01 to $R^2$ for every reported linear outcome.

\begin{table*}[t]
\centering
\scriptsize
\setlength{\tabcolsep}{3pt}
\renewcommand{\arraystretch}{1.08}
\begin{tabular*}{\textwidth}{@{\extracolsep{\fill}}p{3.0cm}ccccc@{}}
\toprule
Term & $\beta$ (SE) & Profile 95\% CI & Penalized LRT $p$ & Holm $p$ & OR [95\% CI] \\
\midrule
GPU count (tenfold increase)
& 0.5380 (0.1883) & [0.1730, 0.8907] & 0.0043 & 0.0085 & 1.713 [1.189, 2.437] \\
Ampere-or-newer/equivalent
& $-0.1115$ (0.2734) & [$-0.6322$, 0.4362] & 0.6833 & 0.6833 & 0.894 [0.531, 1.547] \\
\bottomrule
\end{tabular*}
\caption{Firth penalized logistic regression of paper-award status on reported GPU count and hardware generation. The model uses the same 5,357 papers and 111 award-positive cases as the joint linear probability model. GPU count and the Ampere-or-newer/equivalent indicator are entered simultaneously with the baseline fixed effects and controls. The joint null that both hardware coefficients equal zero is rejected, $\chi^2(2)=8.3178$, $p=0.0156$.}
\label{tab:award-firth-joint}
\end{table*}

\begin{table*}[t]
\centering
\scriptsize
\setlength{\tabcolsep}{3.5pt}
\renewcommand{\arraystretch}{1.06}
\begin{tabular*}{\textwidth}{@{\extracolsep{\fill}}p{3.0cm}p{3.2cm}ccccc@{}}
\toprule
Outcome & Specification & $\beta$ (SE) & 95\% CI & $p$ & $R^2$ controls/full & $\Delta R^2$ \\
\midrule
NLP topic--year percentile
& Common-sample baseline & 0.0313 (0.0118) & [0.0081, 0.0545] & 0.008 & 0.0895/0.0928 & 0.0032 \\
& + pre-publication controls & 0.0274 (0.0121) & [0.0037, 0.0512] & 0.024 & 0.1330/0.1354 & 0.0023 \\
& + public artifact & 0.0265 (0.0121) & [0.0028, 0.0502] & 0.028 & 0.1362/0.1384 & 0.0022 \\
\addlinespace
OpenAlex field-normalized percentile
& Common-sample baseline & 0.0116 (0.0091) & [$-0.0062$, 0.0295] & 0.202 & 0.0909/0.0916 & 0.0008 \\
& + pre-publication controls & 0.0105 (0.0094) & [$-0.0080$, 0.0290] & 0.264 & 0.1172/0.1178 & 0.0006 \\
& + public artifact & 0.0102 (0.0095) & [$-0.0084$, 0.0287] & 0.282 & 0.1179/0.1185 & 0.0006 \\
\addlinespace
$\log(1+\mathrm{citations})$
& Common-sample baseline & 0.1617 (0.0507) & [0.0623, 0.2611] & 0.001 & 0.2109/0.2153 & 0.0044 \\
& + pre-publication controls & 0.1438 (0.0521) & [0.0417, 0.2458] & 0.006 & 0.2455/0.2488 & 0.0033 \\
& + public artifact & 0.1404 (0.0520) & [0.0385, 0.2422] & 0.007 & 0.2477/0.2508 & 0.0031 \\
\bottomrule
\end{tabular*}
\caption{Robustness of aggregate reported GPU-capability estimates to expanded observable controls. All models use the same complete-case sample of 2,077 papers. The coefficient represents the association with a tenfold increase in aggregate reported GPU capability. Incremental $R^2$ compares each full model with a model containing the same controls but excluding reported GPU capability.}
\label{tab:impact-confounding-robustness}
\end{table*}

\subsection{Rare-Event Robustness for Paper Awards}
\label{app:award-firth}

Award status is rare in the strict 2020--2025 sample: 111 of 5,357
papers receive an award label. We therefore re-estimate the joint
GPU-count and hardware-generation model using Firth penalized logistic
regression, with the same analysis sample, fixed effects, and controls
as the joint linear probability model. Profile-likelihood confidence
intervals and penalized likelihood-ratio tests are used for inference.
Holm-adjusted $p$-values account for testing the two hardware terms. The resulting rare-event estimates are reported in
\cref{tab:award-firth-joint}.

As shown in \cref{tab:award-firth-joint}, the rare-event model reproduces the dimension-specific pattern from the
linear probability model. Conditional on hardware generation and the
baseline controls, a tenfold increase in reported GPU count is
associated with 1.713 times the odds of receiving an award
(95\% CI [1.189, 2.437], Holm-adjusted $p=0.0085$). The conditional
association for Ampere-or-newer/equivalent hardware is close to zero and
statistically imprecise. Thus, the award results do not support a
uniform association across hardware dimensions: they identify a
positive association with deployment scale, but not with hardware
generation. Given the small number of award-positive papers and the
observational design, this result is interpreted as a rare-event
robustness association rather than evidence that increasing GPU count
causes award recognition.

\subsection{Robustness to Expanded Observable Controls}
\label{app:impact-expanded-controls}

To assess observable confounding, we re-estimate the citation models on
a common complete-case sample of 2,077 papers. The common-sample
baseline retains the original fixed effects and team-structure
controls. The second specification adds pre-publication measures of
author citation history, team publication experience, institutional
citation visibility, and collaboration structure. A third,
additional specification includes public-artifact availability. We
treat artifact availability separately because it may be jointly
determined with other features of the focal research project rather
than being a strictly pre-publication confounder. The resulting
estimates across these specifications are reported in
\cref{tab:impact-confounding-robustness}.

For the primary NLP topic--year percentile, the estimate declines from
3.13 percentage points in the common-sample baseline to 2.74 percentage
points after adding the pre-publication controls and to 2.65 percentage
points after additionally including artifact availability. The
association remains positive, but the incremental $R^2$ declines from
0.0032 to 0.0023 and 0.0022. The OpenAlex field-normalized estimates
remain small and statistically imprecise. The log-citation estimate is
also attenuated but remains positive. These results indicate that
richer observable controls explain additional variation in citation
impact without making reported GPU capability a substantial source of
incremental explanatory power.

\section{Generalizability Beyond Main-Conference Papers}
\label{app:findings-extension}

\begin{table*}[t]
\centering
\small
\setlength{\tabcolsep}{8pt}
\renewcommand{\arraystretch}{1.08}
\begin{tabular}{lrrr}
\toprule
Characteristic
& Main Conference
& Findings
& Main + Findings \\
\midrule
Total papers
& 13,921
& 9,917
& 23,838 \\

Papers reporting standardized GPU model
& 6,900
& 5,824
& 12,724 \\

Reporting rate (\%)
& 49.6
& 58.7
& 53.4 \\

Papers reporting GPU model + count
& 5,360
& 4,186
& 9,546 \\

Strict reporting rate (\%)
& 38.5
& 42.2
& 40.0 \\

Citation-analysis sample
& 2,194
& 1,620
& 3,814 \\

Median reported GPU count
& 4
& 2
& 3 \\

Median reported GPU capability (TFLOP/s)
& 455.2
& 359.7
& 448.0 \\
\bottomrule
\end{tabular}

\vspace{2pt}
\parbox{0.97\textwidth}{\footnotesize
\textit{Notes:}
Counts and reporting rates are based on papers published during 2020--2025.
The citation-analysis sample contains papers published during 2020--2023 that report both a standardized GPU model and an explicit GPU count.
The median reported GPU count and median reported GPU capability are calculated over the 2020--2025 strict sample.
Reported GPU capability denotes the maximum text-reported GPU configuration capability identified for each paper.
}
\caption{Sample coverage and reported GPU resources in Main Conference and Findings papers.}
\label{tab:findings-coverage}
\end{table*}

\paragraph{Motivation and extension sample.}
The main analyses focus on papers published in the main tracks of ACL, EMNLP, and NAACL. Because selection into a main-conference track may restrict the range of papers represented in the analysis, we assess whether the main findings generalize to the corresponding Findings papers. We apply the same paper-selection, full-text processing, GPU-information extraction, hardware standardization, citation matching, topic classification, and covariate-construction procedures to Findings papers published during 2020--2025. Because conference awards are track-specific recognition outcomes and are not directly comparable across Main Conference and Findings papers, this extension focuses on citation-based measures of scholarly impact.

As shown in \cref{tab:findings-coverage}, the extension adds 9,917 Findings papers to the 13,921 Main Conference papers, producing a combined corpus of 23,838 papers. Findings papers have somewhat higher GPU-reporting rates: 58.7\% report a standardized GPU model and 42.2\% report both a standardized model and an explicit GPU count, compared with 49.6\% and 38.5\%, respectively, among Main Conference papers. At the same time, the median reported GPU count and aggregate capability are lower in Findings. Thus, the Findings sample broadens publication-track coverage while also introducing a meaningfully different distribution of reported computational resources.

\paragraph{Replication of citation-impact associations.}
We next re-estimate the main citation models separately for Main Conference and Findings papers and in their pooled sample. All analyses use the strict 2020--2023 sample and retain the outcome definitions and model specifications used in the main analysis. Reported GPU capability enters each model as
$c_i=\log_{10}(C_i)$, so a one-unit increase corresponds to a tenfold increase in reported GPU capability.

The common linear predictor can be written as
\begin{equation}
\label{eq:findings-common-slope}
\eta_i
=
\beta c_i
+
\mathbf{X}_i^{\prime}\boldsymbol{\gamma}
+
\alpha_{yvt}
+
\delta_q
,
\end{equation}
where $\mathbf{X}_i$ contains team-size and organization-count controls,
$\alpha_{yvt}$ denotes publication-year-by-venue-by-track fixed effects in the pooled models and the corresponding publication-year-by-venue fixed effects in the track-specific models, and $\delta_q$ denotes primary-topic fixed effects. The identity link is used for the citation-percentile, log-citation, and linear-probability specifications, whereas the citation-count model is estimated using Poisson pseudo-maximum likelihood (PPML) with a log link.

For the OLS and linear-probability models, we additionally report the incremental explanatory power of reported GPU capability:
\begin{equation}
\label{eq:findings-incremental-r2}
\Delta R^2
=
R^2_{\mathrm{full}}
-
R^2_{\mathrm{controls}},
\end{equation}
where the full model corresponds to the specification in
\cref{eq:findings-common-slope}, and the controls-only model removes $c_i$ while retaining the same observations, fixed effects, and covariates. Thus, $\Delta R^2$ measures the additional in-sample explanatory power associated with adding reported GPU capability to an otherwise identical model. Ordinary $R^2$ is not defined for the PPML specification and is therefore not reported for the citation-count outcome.

The pooled coefficients in \cref{tab:findings-regressions} are obtained from common-slope models using \cref{eq:findings-common-slope}. To test whether the estimated association differs between publication tracks, we additionally estimate
\begin{equation}
\label{eq:findings-interaction}
\begin{aligned}
\eta_i
={}&
\beta_{\mathrm{M}}c_i
+
\Delta\beta
\left(c_i \times \mathit{Findings}_i\right)
\\
&+
\mathbf{X}_i^{\prime}\boldsymbol{\gamma}
+
\alpha_{yvt}
+
\delta_q
,
\end{aligned}
\end{equation}
where $\beta_{\mathrm{M}}$ is the Main Conference slope and
$\Delta\beta$ is the Findings-minus-Main slope difference. The main effect of the Findings indicator is absorbed by the publication-year-by-venue-by-track fixed effects.

\begin{table*}[t]
\centering
\scriptsize
\setlength{\tabcolsep}{3.8pt}
\renewcommand{\arraystretch}{1.10}
\begin{tabular}{p{4.15cm}cccc}
\toprule
Outcome
& \makecell{Main Conference\\$\beta$ (SE) [$\Delta R^2$]}
& \makecell{Findings\\$\beta$ (SE) [$\Delta R^2$]}
& \makecell{Pooled\\$\beta$ (SE) [$\Delta R^2$]}
& \makecell{Findings $-$ Main\\difference ($p$)} \\
\midrule

NLP topic--year citation percentile
& 0.035 (0.011) [0.0042]
& 0.046 (0.014) [0.0071]
& 0.039 (0.009) [0.0051]
& $+0.011$ (0.551) \\

OpenAlex field-normalized percentile
& 0.013 (0.009) [0.0009]
& 0.040 (0.011) [0.0073]
& 0.024 (0.007) [0.0029]
& $+0.027$ (0.051) \\

$\log(1+\mathrm{citations})$
& 0.172 (0.049) [0.0052]
& 0.207 (0.054) [0.0083]
& 0.185 (0.036) [0.0061]
& $+0.035$ (0.630) \\

Citation count (PPML)
& 0.477 (0.085) [---]
& 0.390 (0.090) [---]
& 0.456 (0.066) [---]
& $-0.088$ (0.481) \\

Top-10\% cited (LPM)
& 0.038 (0.014) [0.0043]
& 0.059 (0.018) [0.0093]
& 0.047 (0.011) [0.0062]
& $+0.020$ (0.372) \\

\midrule
$N$
& 2,194
& 1,620
& 3,814
& 3,814 \\
\bottomrule
\end{tabular}

\vspace{2pt}
\parbox{0.98\textwidth}{\footnotesize
\textit{Notes:}
Cells in the first three estimate columns report $\beta$ (robust SE) [$\Delta R^2$].
All models use the strict 2020--2023 sample and include publication-year-by-venue fixed effects in the track-specific analyses and publication-year-by-venue-by-track fixed effects in the pooled analyses, together with primary-topic fixed effects, team-size controls, and organization-count controls.
Reported GPU capability is entered as $\log_{10}(C_i)$; coefficients therefore correspond to a tenfold increase in reported GPU capability.
Incremental $R^2$ is the difference between the full specification and a controls-only model estimated on the identical sample; the two models differ only in the inclusion of reported GPU capability.
The Main Conference and Findings columns are estimated separately, whereas the pooled column reports the common-slope specification in \cref{eq:findings-common-slope}.
The final column reports the coefficient on the interaction between reported GPU capability and the Findings indicator in \cref{eq:findings-interaction}, followed by its two-sided Wald-test $p$-value in parentheses.
OLS and linear-probability models use HC3 robust standard errors.
The PPML model uses HC0 robust standard errors; ordinary $R^2$ is not defined for PPML and is therefore not reported.
The PPML coefficients imply expected-citation differences of 61.2\%, 47.7\%, and 57.8\% per tenfold increase in reported GPU capability for the Main Conference, Findings, and pooled samples, respectively.
None of the five publication-track differences is statistically significant at the 5\% level, and none remains significant after Holm correction across the five difference tests.
}
\caption{Replication of citation-impact associations and incremental explanatory power across publication tracks.}
\label{tab:findings-regressions}
\end{table*}

\begin{table*}[t]
\centering
\scriptsize
\setlength{\tabcolsep}{4pt}
\renewcommand{\arraystretch}{1.10}
\begin{tabular}{lccccc}
\toprule
Track
& \makecell{High-impact rate among\\high-capability papers}
& \makecell{High-impact rate among\\other papers}
& \makecell{Risk\\ratio}
& \makecell{High-capability share among\\high-impact papers}
& \makecell{High-capability papers\\not high-impact} \\
\midrule

Main Conference
& 14.5\%
& 9.1\%
& 1.59
& 28.6\%
& 85.5\% \\

Findings
& 15.8\%
& 9.0\%
& 1.75
& 30.6\%
& 84.2\% \\

Main + Findings
& 14.7\%
& 9.2\%
& 1.60
& 28.6\%
& 85.3\% \\

\bottomrule
\end{tabular}

\vspace{2pt}
\parbox{0.98\textwidth}{\footnotesize
\textit{Notes:}
High capability denotes the annual top 20\% of papers by text-reported maximum GPU capability.
High impact denotes the top 10\% of papers by citation count within venue-by-publication-year-by-track cells.
The analysis uses the 2020--2023 model-reported GPU sample:
$N=3{,}156$ for Main Conference papers,
$N=2{,}427$ for Findings papers, and
$N=5{,}583$ for the combined sample.
For the pooled row, the annual high-capability threshold is recalculated over the combined sample and is therefore not an arithmetic aggregation of the two track-specific rows.
Risk ratios are descriptive associations and should not be interpreted as causal effects.
}
\caption{Overlap between high reported GPU capability and high citation impact across publication tracks.}
\label{tab:findings-overlap}
\end{table*}

The estimated association is positive across all five citation outcomes in both publication tracks. For the primary NLP topic--year citation percentile, a tenfold increase in reported GPU capability is associated with increases of 3.5 percentage points among Main Conference papers and 4.6 percentage points among Findings papers. The corresponding pooled estimate is 3.9 percentage points. Adding reported GPU capability increases $R^2$ by 0.0042 in the Main Conference sample, 0.0071 in the Findings sample, and 0.0051 in the pooled sample. The Findings-minus-Main slope difference is small and statistically imprecise
($\Delta\beta=0.011$, $p=0.551$).

Across the OLS and linear-probability outcomes, the incremental $R^2$ values range from 0.0009 to 0.0093. They are numerically larger in the Findings sample for each outcome, but remain below 0.01 in both publication tracks. Reported GPU capability therefore contributes additional explanatory power beyond the fixed effects and observed controls, but the magnitude of that contribution remains modest. These numerical differences in incremental $R^2$ are descriptive and should not be interpreted as formal tests of publication-track heterogeneity.

The largest estimated slope difference occurs for the OpenAlex field-normalized citation percentile, for which the Findings coefficient exceeds the Main Conference coefficient by 2.7 percentage points. However, this difference does not reach the conventional 5\% significance threshold ($p=0.051$) and does not remain significant after correction for the five track-comparison tests. Track differences are also statistically insignificant for the NLP topic--year citation percentile, log citations, citation counts, and the probability of being among the top 10\% most cited papers. The positive but limited association between reported GPU capability and citation impact is therefore not confined to papers selected into the main-conference tracks.

\paragraph{Overlap between high capability and high impact.}
We also replicate the descriptive overlap analysis to assess whether the central pattern of positive but limited alignment between reported GPU capability and citation impact extends to Findings papers. High capability denotes the annual top 20\% of papers by text-reported maximum GPU capability. High impact denotes the top 10\% of papers by citation count within venue-by-publication-year-by-track cells.

This analysis uses the broader 2020--2023 model-reported GPU sample, matching the sample definition used for the corresponding analysis in the main paper. It contains 3,156 Main Conference papers, 2,427 Findings papers, and 5,583 papers in the combined sample. For the pooled row, the annual high-capability threshold is recalculated over the combined Main Conference and Findings sample rather than obtained by aggregating the two track-specific classifications.

As shown in \cref{tab:findings-overlap}, the overlap results are highly similar across publication tracks. Among Main Conference papers, 14.5\% of high-capability papers are highly cited, compared with 9.1\% of other papers, corresponding to a descriptive risk ratio of 1.59. Among Findings papers, the corresponding rates are 15.8\% and 9.0\%, yielding a risk ratio of 1.75. The pooled risk ratio is 1.60.

Nevertheless, greater reported GPU capability does not ensure high citation impact in either track. Among high-capability papers, 85.5\% of Main Conference papers and 84.2\% of Findings papers are not in the top 10\% of the citation distribution. Conversely, high-capability papers account for only 28.6\% of highly cited Main Conference papers and 30.6\% of highly cited Findings papers. The broader publication-track analysis therefore reproduces the main conclusion: reported GPU capability is positively associated with citation impact, but the alignment is limited, and high reported capability is neither sufficient nor necessary for high citation impact.

\end{document}